\documentclass{article}

\usepackage[letterpaper,text={5.5in,9in},centering]{geometry}
\usepackage{times}
\usepackage{mathptmx}
\usepackage{fancyhdr}
\usepackage{titlesec}

\usepackage{iclr2027_conference,times}
\iclrfinalcopy

\titlespacing*{\paragraph}{0pt}{1.5ex plus .4ex minus .2ex}{1em}

\usepackage{natbib}
\usepackage{amsmath,amssymb}
\usepackage{booktabs}
\usepackage{xcolor}
\usepackage{graphicx}
\usepackage{enumitem}
\usepackage[colorlinks=true,allcolors=blue]{hyperref}
\usepackage{url}
\usepackage{caption}
\newcommand{\rgs}{\textsc{rgs}}
\newcommand{\caa}{\textsc{caa}}
\newcommand{\cccaa}{class-conditional \caa}

\title{When Should an LLM Steering Edit Depend on the Input?
}

\author{Rachael Harkness \& Konrad Rawlik \\
Baillie Gifford Pandemic Science Hub, Centre for Inflammation Research \\
The University of Edinburgh \\
\texttt{\{rachael.harkness,konrad.rawlik\}@ed.ac.uk}
}

\date{}

\begin{document}

\maketitle

\begin{abstract}
An activation steering edit should depend on the input exactly when the destination varies
\emph{inside} the class a query requests, and that is measurable before any rule is fitted or applied.
Contrastive Activation Addition (\caa) adds one fixed direction to every input, the
difference of the class means. Conditioning it on the requested class accounts for the
between-class variation in the required displacement and leaves the within-class
variation untouched, which no rule that reads only the source state can express. We study 
the impact of increasing input-dependence in the steering edit, moving from a difference of means 
to a destination retrieved per query --- the mean of the $k$ stored states nearest the query, inside 
the requested class --- in which the two baselines are endpoints: at $k$ equal to the class size the  retrieved rule \emph{is} the
class-conditional difference of means. We measure input-dependent steering on six substitution corpora that vary in whether the edit changes a word or a
clause, and on an external benchmark, and find that input-dependent steering outperforms fixed-direction steering when the destination varies within the class.
Retrieval wins at every layer of every substitution corpus in two models, and the level of input-dependence needed governs how to apply it: 
word-level edits are best made in the first tenth of the network from a few neighbours, clause-level edits at any depth from the class mean. 
Where the class already determines the answer the class mean
suffices and retrieval only matches it, though a fixed direction still fails there. Four quantities
computed from the states alone identify the regime in which the edit should be applied.
\end{abstract}

\newpage

\section{Introduction}

Activation steering perturbs hidden states with a vector chosen to induce a behaviour. \caa\
collects states from prompts exhibiting the behaviour and from prompts that do not, and adds
the difference of their means to every input. This assumes one displacement describes the
edit for every input, and the assumption fails whenever the required edit depends on what the
input says: steering a sentence toward the next weekday asks something different of
\emph{Monday} and \emph{Thursday}, where steering a concept from male to female
(\emph{king} to \emph{queen}) does not.

\begin{figure}[h!]
\centering
\includegraphics[width=0.86\linewidth]{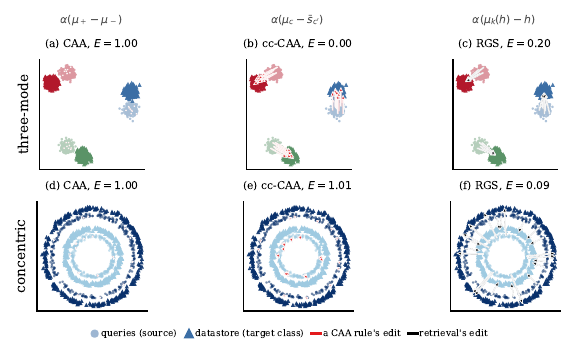}
\caption{A fixed direction fails where the required displacements disagree, and a per query destination recovers the task. Three rules on two generated constructions in two
dimensions; $E$ is reference error, $1$ being the score for leaving the
input alone.
\textsc{three-mode} (a--c) rotates the offsets between classes, so \caa's vector is zero by
construction while the same difference of means taken inside a class is exact (\cccaa);
\textsc{concentric} (d--f) sends each query outward at its own angle, so the displacements
cancel within a class too and neither fixed rule has anything to estimate. Only a \rgs\
destination survives both (\S\ref{app:synthetic} gives more detail).}
\label{fig:method}
\end{figure}

Two repairs are available, and they are the same rule at different settings. Conditioning the
difference of means on the class a query is steered toward gives one vector per class rather
than one per corpus. Retrieving instead gives a destination per query: search a bank of
states that already exhibit the target behaviour and steer toward the mean of the $k$
nearest. \S\ref{sec:endpoints} shows these are endpoints of one family --- at $k$ equal to
the class size the retrieved rule \emph{is} the class-conditional difference of means --- so
$k$ controls the locality of the edit and therefore how closely the retrieved destination matches the ideal per-query direction; 
this characterises the trade-off between global and local edits and defines the difference between class-conditional and per-query retrieval. Figure \ref{fig:method} displays the effects of per-query retrieval.

This trade-off is task-dependent, and can be characterised quantitatively before any rule is fitted. 
Each rule has the same form, one direction per query times a scalar chosen on held-out data,
 so each can be scored in advance by how well its directions agree with the edits the task requires. 
We compute four such quantities and measure them against what the rules go on to score.

We evaluate on five families: synthetic constructions whose correct answers are set rather than
measured, two token corpora, six substitution corpora built to vary the level of the concept
an edit changes (word to clause), two A/B behavioural benchmarks, 
and two external benchmarks run under their own protocols. 
Throughout we compare \rgs\ against \caa, \cccaa\ and an affine map fitted on
the query-to-target correspondence none of the unpaired rules receive, and we ask what 
governs the choice between them: the level of the concept, the coverage of the bank, the
depth at which the edit is made, and whether the destination varies inside the class a query
requests.

\section{Related work}
\label{sec:related}

\paragraph{Steering by a fixed direction.} Most activation steering adds one vector to every
input: activation addition \citep{turner2023} and Contrastive Activation Addition
\citep{rimsky2023} estimate it from contrastive prompt pairs, inference-time intervention
\citep{li2023} fits per-head directions to a probe, representation engineering
\citep{zou2023} builds reading and control vectors from the same contrast, and task
arithmetic \citep{ilharco2023} composes fixed directions in weight space. All share the
assumption examined here --- that one displacement describes the edit for every input --- and
all are evaluated by class-level outcomes, which a rule collapsing every query onto the class
centroid satisfies exactly. Evidence that some model features are not linearly organised
\citep{engels2024} suggests the assumption is not universal, without saying which tasks
violate it.

\paragraph{Conditioning the intervention on the input.} Function vectors \citep{todd2023}
condition on the \emph{task}, applying one vector per task rather than per query. Retrieval over a datastore of hidden states is established as a way of conditioning a model's \emph{output} on nearest neighbours \citep{khandelwal2020,khandelwal2021,shi2022,huang2023},
which interpolate the output distribution with a neighbour-weighted one. We use the same machinery for a different purpose: the neighbourhood supplies a steering \emph{destination} in state space, so the intervention remains a single additive edit at one layer and composes with the fixed-direction methods above.

\paragraph{Our contributions.}

\begin{enumerate}[label=(\roman*),leftmargin=1.6em,itemsep=3pt,topsep=3pt,parsep=0pt]

\item We introduce \rgs, a rule for input-dependent steering of which \cccaa\ is the $k=|\mathcal{D}_c|$ endpoint (\S\ref{sec:endpoints}).

\item We evaluate \rgs\ against alternative steering methods on a variety of corpora to identify the successes and failures relative to fixed steering rules.  

\item We identify that concept level (words vs. phrases) organises where in the model the edit is best made, and how local the rule should be (\S\ref{sec:results}).

\item We define four diagnostic quantities computed from the task alone which predict the success of each rule and propose their use as a screening tool for input-dependent steering (\S\ref{sec:screen}).

\end{enumerate}

\section{Methods}
\label{sec:method}

We compare input-dependent rules against fixed rules. Both apply the same general method i.e., a displacement is added to a source state; they differ only in how much of the data bank the
destination is read from --- the whole corpus (\caa), one class (\cccaa), or the $k$ entries nearest the query (\rgs). For \rgs\ and \cccaa, $k$ names which rule is running, and choosing $k$ is defines the locality the edit should be.

\subsection{Steering rules}

Let $\mathcal{D}$ be a bank of target states and $\mathcal{S}$ a set of source states.
Writing each rule as the displacement it adds,
\begin{equation}
\delta_{\caa}(h)=\alpha\bigl(\operatorname{mean}(\mathcal{D})-\operatorname{mean}(\mathcal{S})\bigr),
\qquad
\delta_{\rgs}(h)=\alpha\bigl(\mu_k(h)-h\bigr),
\quad
\mu_k(h)=\tfrac{1}{k}\!\!\sum_{x\in N_k(h)}\!\!x,
\label{eq:rules}
\end{equation}
where $N_k(h)$ are the $k$ elements of $\mathcal{D}$ nearest $h$ using a Euclidean distance. 
The search is limited to target states, entries remain unlinked to any source. $\alpha$ is a scalar value that weights the displacement. Both $\alpha$ and $k$ are selected on a validation split and reported unchanged on test.

With $C$ for the number of classes, $\mathcal{D}_c$ for the entries labelled $c$ and
$\mathcal{S}_{c'}$ for the sources steered toward it, \cccaa\ is equation \eqref{eq:rules}'s difference of means taken inside a class,
$\alpha(\operatorname{mean}(\mathcal{D}_c)-\operatorname{mean}(\mathcal{S}_{c'}))$, which gives one
vector per class rather than one for the corpus. We additionally consider a fitted affine map (defined in \S\ref{app:affine}). The fitted affine map requires aligned pairs and is reported throughout as a ceiling on the comparison . 

\subsubsection{\rgs\ and \cccaa\ coincide at $k=|\mathcal{D}_c|$}
\label{sec:endpoints}

Restricting the search to class $c$ and taking $k=|\mathcal{D}_c|$ gives
$\mu_k(h)=\operatorname{mean}(\mathcal{D}_c)$, so
\begin{equation}
\delta_{\rgs}(h)\big|_{k=|\mathcal{D}_c|}=\alpha\bigl(\operatorname{mean}(\mathcal{D}_c)-h\bigr),
\qquad
\delta_{\cccaa}(h)=\alpha\bigl(\operatorname{mean}(\mathcal{D}_c)-\operatorname{mean}(\mathcal{S}_{c'})\bigr),
\label{eq:endpoint}
\end{equation}
and since $\mathbb{E}[h]=\operatorname{mean}(\mathcal{S}_{c'})$ the two coincide in expectation: \cccaa\ aims at the target class centroid from the source class centroid, \rgs\ from the query. 
At $k=1$ it transplants a single stored state, at the class ceiling it \emph{is} \cccaa\, so the selected $k$ reports which rule won.

\subsection{Corpora}
\label{sec:corpora}

Five families: six substitution corpora, two token corpora, two A/B behavioural benchmarks,
generated constructions with set geometries, and an external
benchmark that runs under its own protocol (\S\ref{sec:corpora-external}). All but the A/B benchmarks supply a \emph{per-query reference}, i.e. the state the model would have produced had the correct edit already been made, which is what makes reference error and per-item identity computable (\S\ref{sec:metrics}). Corpora are split by frames, so no frame appears in both a fitted rule's training pairs and its evaluation queries, and steering results are four-fold cross-validated. Reference errors are normalised by the whole corpus's required step rather than each fold's, so the unsteered score reads $1.000\pm0.073$ rather than exactly $1$ (\S\ref{app:cv}).

\subsubsection{Substitution corpora}

Six corpora that vary one property while holding everything else fixed (Table~\ref{tab:corpora}, \S\ref{app:examples}).
Each corpora pairs a source span with a destination span in every class, renders both in the
\emph{same} sentence frame and reads the last-token state, so a source and its reference
differ by a single property, e.g., the lexical or propositional decision, and by nothing else (frames, rendered
examples and the rules keeping class vocabularies equal). Four are
word-level: \textsc{lexicon} maps an English noun to its French, German or Spanish equivalent
(\emph{water} $\rightarrow$ \emph{eau}), so source and destination vocabularies are disjoint;
\textsc{tense} maps an irregular verb to its past, progressive or third-person form; \textsc{register} maps a neutral word to a formal or casual
synonym; and \textsc{casing} upper-cases or capitalises a
word and is the theoretical control, every query needing the same edit. Two are clause-level:
\textsc{sentiment} replaces a neutral clause with a positive or negative one and is the only
corpus whose source and reference are two different clauses rather than one clause with a
span replaced, and \textsc{certainty} asserts or hedges a proposition with both renderings
\emph{ending on the same word}. The size of each corpus is given in Table \ref{tab:corpora} in \S\ref{app:examples}.

\subsubsection{Token corpora}
Two corpora that define two steering tasks by varying a single token in the frame.  Weekday advancement steers a sentence naming one day toward the next (``\dots moved to
Monday'' $\rightarrow$ ``\dots to Tuesday''), over $1200$ templates and seven days. Source
and target sets are the \emph{same states permuted}, so
$\operatorname{mean}(\mathcal{D})=\operatorname{mean}(\mathcal{S})$ and \caa's vector is zero
by identity rather than estimation error; the day centroids are near-equidistant rather than
ringed, so each day requires a different step (\S\ref{app:weekday-geometry}).
Country-to-capital steers toward a capital (``\dots held in France'' $\rightarrow$ ``\dots in
Paris''), over $20$ country/capital pairs, $40$ frames and $12$ subjects.

\subsubsection{Behavioural benchmarks}
\label{sec:corpora-behavioural}

Two A/B corpora from the model-written evaluations of \citet{perez2022}, sycophancy and
corrigibility, each presenting a question with two answers, one of which exhibits the
behaviour. They supply no per-query reference --- the two states of an item differ mostly in
which answer is selected --- so reference error and identity are undefined and we score them
by forced choice (\S\ref{sec:metrics}).

\subsubsection{CounterFact benchmark}
\label{sec:corpora-external}

We use one external benchmark and score it the way its own literature scores it. \textsc{CounterFact} \citep{meng2022} asks a subject-relation prompt to end in a new object; the bank is the benchmark's own attribute prompts, states of other subjects that already carry that object, and the score is its efficacy measure --- the fraction of cases where the edited model prefers the new object to the true one. Protocol in full in \S\ref{app:counterfact}.

\subsubsection{Generated constructions}
\label{sec:corpora-generated}

We generate four two-dimensional datasets with set geometric arrangements. Each dataset is constructed to demonstrate the failure cases of fixed steering rules, so every rule's correct score is known before it runs. \caa\ adds the difference of class means to every query, so the only part of the required edit it can supply is the part every query shares: with displacements $d_i = r_i - h_i$, its vector is $\mathbb{E}[d]$ and what it necessarily leaves is their spread about that mean. Each construction removes that shared part in a different way.

\textsc{three-mode} rotates each of three clusters' offsets, so the required displacements cancel across classes and \caa\ is left with a zero vector, while the same difference of means taken inside a class is exact. We include a \textsc{shared-offset} control that varies globality alone to make the task solvable by \caa. \textsc{concentric} puts the classes on rings and sends each query radially outward at its own angle; radial vectors average to zero around a ring, so the displacements cancel \emph{within} a class as well and \cccaa\ has nothing to estimate either, and the two ring ratios cannot be served by one linear map. We include a \textsc{hierarchical} construction that places every class inside every cluster and gives each cluster its own rotation, so the displacement a class requires depends on which cluster the query sits in: the linear part of the required map is shared across clusters but the intercept is not. Described in detail in Figure \ref{fig:synthetic} in \S\ref{app:synthetic}.

\subsection{Diagnostics}
\label{sec:diagnostics}

Before either rule is run we compute four diagnostic quantities that describe a steering task: the standardised displacement $D=\sqrt{\mathbb{E}\lVert
d\rVert^2}/\sqrt{\mathbb{E}\lVert x-\bar x\rVert^2}$, how far the state must move in units of
the states' own spread; the global shared fraction $g=\lVert\mathbb{E}[d]\rVert/
\mathbb{E}\lVert d\rVert$, whether the per-sample displacements $d_i=r_i-h_i$ point the same way; the
residual $R=\sqrt{\mathbb{E}\lVert d-\bar d\rVert^{2}}/\sqrt{\mathbb{E}\lVert x-\bar
x\rVert^{2}}$, the portion of the edit no single vector supplies; and the
retrieval alignment $A$, agreement between optimal retrieved displacement and true displacement, reported as $1-A$ to keep to the same units as $E$. $D$, $g$ and $R$
need the pairs alone; $A$ also needs class labels and a bank, so every one of them asks for more supervision than the rule it screens for. 
Definitions in full are in \S\ref{app:diagnostics}, and the sample size each requires, six pairs for $R$, thousands for $g$, and hundreds for $A$, in \S\ref{app:samples}.

\subsection{Evaluation}
\label{sec:metrics}

Six metrics are reported, in two groups; definitions in full, with the estimator choices, are in \S\ref{app:metrics}. In state space, measured against the ground truth reference: 
\emph{reference error} $E$, the distance from a steered state to its own reference in units of the corpus's mean required step, 
so $E=1$ is the score for no change to the input; \emph{per-item identity},
the fraction of steered states whose nearest reference among all test references is its own;
and \emph{destination rate}, the fraction whose nearest class centroid is its target.
On the decoded text: \emph{span match}, whether the continuation contains
the destination span; \emph{forced choice}, the share of probability on the requested
destination against that concept's destinations in the other classes, class-balanced so a
fixed preference scores exactly $1/C$ (the chance level for a random guess); and \emph{fluency}, the log-probability of the steered
continuation under the unsteered model, read against that model's own greedy continuation
with only the source span rewritten.

A query's \emph{source} is a concept rendered in a frame; its \emph{reference} is that same
frame with the requested class's destination substituted in. The source is therefore fixed by
the concept alone and the reference by the concept \emph{and} the class, so the $C$ queries
that steer one concept to its $C$ destinations have $C$ distinct references but one shared
source state $h$.
Any rule whose edit is a function of the source state alone --- \caa, the fitted affine map,
any single vector added at a fixed $\alpha$ --- is never told which class is requested, so all
$C$ receive the same edit and land on the same steered state, and one state is nearest to at
most one of those $C$ references (visualised in Figure~\ref{fig:method}). Per-item identity
equation \eqref{eq:identity} is therefore capped for such a rule at exactly $1/C$, by construction
rather than by estimation error. A rule that reads the requested class, \rgs\ among others,
is not bound. Not every metric is defined on every corpus, since a state-space metric needs a per-query
reference and span match needs a single-word destination; Table~\ref{tab:metricmap} states
which is reported where.

\begin{table}[h]
\caption{Metrics by corpus family. \checkmark: reported here; ---:
not. The A/B and clause-level absences are structural (no per-query reference; span match
needs a single-word destination); the token ones are choices.}
\label{tab:metricmap}
\centering\small
\begin{tabular}{lcccc}
\toprule
& \multicolumn{2}{c}{Substitution} & & \\
\cmidrule(lr){2-3}
Metric & word & clause & Token & A/B \\
\midrule
Reference error $E$ \eqref{eq:referror} & \checkmark & \checkmark & \checkmark & --- \\
Per-item identity \eqref{eq:identity} & \checkmark & \checkmark & --- & --- \\
Destination rate \eqref{eq:destination} & \checkmark & \checkmark & \checkmark & --- \\
Span match & \checkmark & --- & --- & --- \\
Forced choice & \checkmark & \checkmark & --- & \checkmark \\
Fluency, span-swapped & \checkmark & \checkmark & --- & --- \\
\bottomrule
\end{tabular}
\end{table}

\section{Results}
\label{sec:results}

We establish where an input-dependent destination beats a fixed one, and what sets the margin of improvement. We consider performance in: generated constructions with known ground truths (\S\ref{sec:results-synthetic}); substitution tasks which vary in the level of substitution required, from token-level (\S\ref{sec:ladder}) to clause-level (\S\ref{sec:level}); multiple choice (A/B) benchmarks for model behaviour (\S\ref{sec:failure}); and an external benchmark for counterfactual retrieval (\S\ref{sec:icl}). We then identify what bounds that result by evaluating a set of conditions, including, model depth, data bank coverage (\S\ref{sec:coverage}), and answer formatting (i.e., multiple choice).

\subsection{Generated constructions}
\label{sec:results-synthetic}

Each construction is generated to demonstrate the failure case of each fixed rule: \textsc{three-mode}
cancels the required displacements across classes and so defeats \caa, \textsc{concentric}
cancels them within a class and so defeats \cccaa, and \textsc{hierarchical} makes
the edit depend on a cluster the class label does not name and so defeats the affine map. The
geometry is set rather than measured, so each rule's correct score is known before it runs. Results are presented in Table \ref{tab:synthetic} and Figure \ref{fig:synthetic} in \S\ref{app:synthetic}). On \textsc{three-mode}, \textsc{concentric} and \textsc{hierarchical}, \rgs\ reaches $0.200$, $0.081$ and $0.155$ reference error against \caa's $1.000$, $1.001$ and $1.000$; on the \textsc{concentric} and \textsc{hierarchical} geometries \rgs\ also beats \cccaa\ ($1.008$, $1.010$) and the paired affine
map ($0.826$, $0.973$). The control isolates the cause: the same clusters with one shared
offset in place of rotated ones restore \caa\ to $0.012$, so it fails because the required
displacements disagree (Figure~\ref{fig:method}). The constructions also test the screen directly: $A$
anticipates successfully what \rgs\ scores to within $0.020$ on all four (\S\ref{app:synthetic}).

\subsection{Token corpora}
\label{sec:ladder}

On the two token corpora \rgs\ beats \cccaa\ on reference error at matched class information --- $0.307$ against
$0.326$ on weekday and $0.408$ against $0.524$ on capitals, in reference error --- while the affine map (which is given aligned pairs) wins  at $0.195$ and $0.292$. \caa\ reads $1.000$ and $0.778$ (Table~\ref{tab:token} in
\S\ref{app:token}).

Weekday is the sharpest case: the corpus
permutes one set of states, so the two centroids coincide. The seven day centroids are near-equidistant rather than ringed
(\S\ref{app:weekday-geometry}), and a permutation of equidistant points has no preferred
direction. This explains why \caa\ scores equivalent to the no steering value while the affine map reaches $0.195$.  For this geometry, \rgs\ improves on the affine map by $0.112$ despite receiving no pairs at all.

\subsection{Substitution corpora}
\label{sec:level}

\begin{figure}
    \centering
    \includegraphics[width=\linewidth]{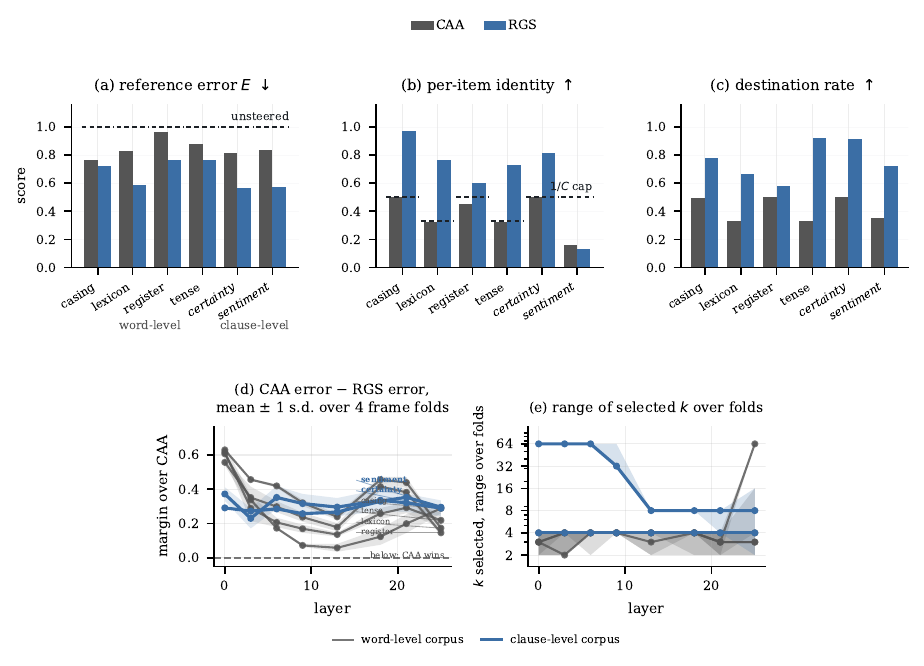}
    \caption{\caa\ against \rgs\ on the six substitution corpora at layer $13$ of Gemma-2-2B-it, grouped by concept level with clause-level in  italics. Dashed lines are equivalent to no steering in (a) reference error and the $1/C$ cap in (b) per-item identity; (c) is the destination rate \eqref{eq:destination}.  Table~\ref{tab:lexical-direct} in \S\ref{app:lexical-direct} gives the same cells numerically. (d) The margin over \caa\ with $\pm1$ s.d.\ bands, (e) range of optimal \rgs $k$ values over 4 cross validation folds. Arrows indicate the direction of improvement.
}
    \label{fig:lexical-direct-cv}
\end{figure}

On the substitution corpora \rgs\ beats \caa\ on every corpus and at every layer of the model where steering is applied,
 narrowest on the \textsc{casing} control at layer $13$ ($0.059\pm0.02$ in reference
error, Figure~\ref{fig:lexical-direct-cv}d) --- the corpus whose required displacements agree most
closely ($g=0.65$ against $0.31$--$0.56$ elsewhere) and where \caa\ is correspondingly at
its strongest.  We find that retrieval consistently improves performance across all substitution corpora, 
with the magnitude of improvement varying according to the level of the concept being edited: a single word in
the four word-level corpora, a whole clause in \textsc{sentiment} and \textsc{certainty}
(\S\ref{sec:corpora}).

Figure~\ref{fig:lexical-direct-cv} gives the comparison at one operating point, layer $13$ with
$\alpha$ and $k$ selected on validation. On reference error \rgs\ wins on all six corpora
and the level does not order the result: the margin over \caa\ is $0.05$ on the control,
$0.12$ to $0.24$ on the other word-level corpora and $0.25$ to $0.26$ on the clause-level
ones, so a clause edit is no harder for retrieval to aim at than a word edit. On per-item identity the two rules
differ in kind rather than degree, \caa\ sitting at or below the $1/C$ cap on every corpus
while \rgs\ clears it on five of six (see \S\ref{app:lexical-direct}). The level also sets where in the model depth the edit is best made: all
four word-level corpora reach their lowest error at layer $0$ and rise steeply out of it,
varying by $0.29$ to $0.45$ across model depth, while the two clause-level corpora vary by
$0.11$ and $0.15$ and \textsc{sentiment}'s and \textsc{clause}'s best layer is $21$ of $26$. We find that the general shape holds where experiments are repeated in Llama-2-7b-chat (\S\ref{app:models}).

By \eqref{eq:endpoint} the selected $k$ reports which rule won, and the depth at which
retrieval becomes local rises with the level: word-level corpora select $k=1$ at the
embedding and loosen to $3$--$6$ through the layers, while \textsc{sentiment} takes the class
ceiling there --- the rule that wins \emph{is} \cccaa\ --- and becomes local only from
layer 13. That locality is worth most of the margin: replacing the selected $k$ with the
class ceiling raises reference error at every layer of every corpus except where the sweep
takes the ceiling anyway, costing $0.44$ to $0.59$ on the four corpora local at layer $0$
(\S\ref{app:ksweep}). We therefore read $k$ as a regime for input-dependent locality rather than a value, and the
regimes are stable across folds (Figure~\ref{fig:lexical-direct-cv}e).

Per-item identity separates the rules far more than reference error does --- \rgs's margin
over \caa\ on \textsc{casing} is $0.467$ in identity against $0.059$ in error --- and Table~\ref{tab:identity} puts both capped
rules beside \rgs\ over four folds. The two fixed rules sit at or below their ceiling on every corpus,
 \caa\ reading $0.310$ to
$0.500$ and the affine map $0.313$ to $0.500$. \rgs, given the requested class as well as the source state, 
clears the ceiling on five corpora, and on four of them roughly doubles what any fixed rule can reach --- $0.967$ and
$0.915$ against a cap of $0.50$, $0.709$ and $0.686$ against $0.33$ --- beating the affine
map by $0.107$ to $0.467$. \textsc{register} is the narrow case, clearing the cap by $0.060$ against a fold
spread of $0.053$; \textsc{sentiment} is the one loss, \rgs\ reading $0.127$ where \caa\
reads $0.123$, so conditioning on the input buys nothing here. Concept level does not order
those gaps --- the two largest margins are the word-level \textsc{casing} control and the
clause-level \textsc{certainty}, and what separates \textsc{certainty} from
\textsc{sentiment} is the span: \textsc{sentiment}
rewrites a whole clause rather than replacing a phrase, so the bank holds no destination for
a particular query, which is reflected in its selected $k$ of $8$ (against $4$ elsewhere), here the neighbourhood is
widening toward a class mean that cannot distinguish the queries it steers.

\begin{table}[h]
\centering
\resizebox{\ifdim\width>\linewidth\linewidth\else\width\fi}{!}{%
\begin{tabular}{llccccr}
\toprule
Corpus & Level & Cap $1/C$ & \caa & Affine map & \rgs & \rgs\ $-$ \caa \\
\midrule
\textsc{casing} & word & $0.50$ & $0.500\pm0.000$ & $0.500\pm0.000$ & $\mathbf{0.967\pm0.031}$ & $+0.467$ \\
\textsc{certainty} & clause & $0.50$ & $0.492\pm0.017$ & $0.475\pm0.022$ & $\mathbf{0.915\pm0.054}$ & $+0.423$ \\
\textsc{tense} & word & $0.33$ & $0.326\pm0.007$ & $0.313\pm0.011$ & $\mathbf{0.709\pm0.080}$ & $+0.383$ \\
\textsc{lexicon} & word & $0.33$ & $0.310\pm0.015$ & $0.321\pm0.011$ & $\mathbf{0.686\pm0.068}$ & $+0.376$ \\
\textsc{register} & word & $0.50$ & $0.446\pm0.020$ & $0.452\pm0.040$ & $\mathbf{0.560\pm0.053}$ & $+0.113$ \\
\textsc{sentiment} & clause & $0.50$ & $0.123\pm0.014$ & $\mathbf{0.340\pm0.056}$ & $0.127\pm0.030$ & $+0.004$ \\
\bottomrule
\end{tabular}}
\caption{Per-item identity at layer 13, mean $\pm$ s.d.\ over four frame folds, ordered by the last column. \caa\ and the affine map read the state alone and are capped at $1/C$ (\S\ref{sec:metrics}), the affine map reaching the cap on \textsc{casing} and never exceeding it.}
\label{tab:identity}
\end{table}

\subsection{Decoded text}
\label{sec:propagation}

When decoding continuations  ($n=60$), span match collapses
with model depth on every corpus while the forced choice does not: \rgs\ holds $0.786$ and $0.784$
at layer 21 on the two three-class corpora against chance $0.333$, so the edit reaches the
output distribution at every depth and ceases to be the greedy argmax past the shallow
layers. \caa\ sits at chance on every corpus at every layer. Span match is undefined at
clause level, reading $0$ there for every rule including those that changed the output, which
is why the forced choice is reported beside it (\S\ref{app:propagation}).

\subsection{Coverage}
\label{sec:coverage}

Retrieval draws on a bank containing the destination in other frames (described in detail in \S\ref{app:examples}). Partitioning the vocabulary so the bank holds one half of the concepts and the queries ask for the other (Table~\ref{tab:coverage}), \rgs\'s margin over the affine map closes to within $0.04$ on the word-level corpora and \textsc{certainty}. The advantage of this family is therefore largely dependent on coverage, and where coverage is insufficient on four of the six corpora the failure mode is merely degradation to \cccaa. Overall, we observe that input-dependent rules outperform fixed rules despite the limitations placed on the bank. 

Under disjoint coverage, on \textsc{certainty} and \textsc{casing} \rgs\ maintains locality and this is where \caa\ wins. \caa\ reads reference errors $0.820$ against \rgs\'s $0.901$ and $0.740$ against $0.790$. Therefore, where the edit does not require a stored destination, a difference of means is the better rule. For \textsc{casing} the edit transforms the word rather than looking it up, which is why it is the only corpus in which \rgs\ maintains it's per-item identity score under reduced coverage ($0.97\rightarrow0.92$). In \textsc{sentiment} the affine map degrades in reference error from $0.430$ to $0.639$ against \rgs\'s $0.573$ to $0.603$, so here the \emph{map's} advantage depends on the vocabulary.  \rgs\ is stable there because it is already reading the class: it selects $k=64$ and gains nothing in identity over \caa\ ($0.127$ against $0.123$), so the partition has nothing to take away.

\begin{table}[h]
\centering
\resizebox{\ifdim\width>\linewidth\linewidth\else\width\fi}{!}{%
\begin{tabular}{lrrrrrrcrr}
\toprule
& \multicolumn{2}{c}{\rgs} & \multicolumn{2}{c}{\caa} & \multicolumn{2}{c}{Affine map} & & \multicolumn{2}{c}{\rgs\ identity} \\
\cmidrule(lr){2-3}\cmidrule(lr){4-5}\cmidrule(lr){6-7}\cmidrule(lr){9-10}
Corpus & full & disj. & full & disj. & full & disj. & $k$ disj. & full & disj. \\
\midrule
\textsc{lexicon} & $0.585$ & $\mathbf{0.768}$ & $0.826$ & $0.858$ & $0.672$ & $0.780$ & $64$ & $0.76$ & $0.44$ \\
\textsc{tense} & $0.762$ & $\mathbf{0.831}$ & $0.882$ & $0.880$ & $0.826$ & $0.872$ & $32$ & $0.73$ & $0.65$ \\
\textsc{register} & $0.767$ & $\mathbf{0.883}$ & $0.962$ & $0.984$ & $0.779$ & $0.886$ & $64$ & $0.60$ & $0.36$ \\
\textsc{sentiment} & $0.573$ & $\mathbf{0.603}$ & $0.834$ & $0.874$ & $0.430$ & $0.639$ & $64$ & $0.13$ & $0.13$ \\
\textsc{certainty} & $0.565$ & $0.901$ & $0.818$ & $\mathbf{0.820}$ & $0.678$ & $0.904$ & $32$ & $0.82$ & $0.60$ \\
\textsc{casing} & $0.721$ & $0.790$ & $0.768$ & $\mathbf{0.740}$ & $0.766$ & $0.822$ & $16$ & $0.97$ & $0.92$ \\
\bottomrule
\end{tabular}}
\caption{Reference error at layer 13 under full coverage and with the vocabulary partitioned so the bank holds one half of the concepts and the queries ask for the other, leaving no query's own destination stored. Lower is better; $k$ and identity are retrieval's. \rgs\ keeps a margin over \caa\ on four corpora under the split and loses it on \textsc{certainty} and \textsc{casing}, where a difference of means becomes the better rule.}
\label{tab:coverage}
\end{table}

\subsection{A/B benchmarks}
\label{sec:failure}

On A/B behavioural benchmarks in Llama-2-7b-chat, scored by the share of probability on the answer exhibiting the behaviour, \rgs\ wins on \textsc{sycophancy} ($0.496$ against \caa's $0.543$, from $0.570$ unsteered) and loses on \textsc{corrigibility} ($0.288$ against $0.230$, from $0.392$) --- the one case where a difference of means beats an input-conditional target. 

Here, the answer format is the primary cause. An item's two states differ mostly in which option is selected, and that token-level flip dominates the displacement; pooled across $k$ items the flip directions cancel. Retrieval inherits the problem: its class-restricted neighbourhood holds states sharing the behaviour label but differing in answer letter, so its mean points along the letter axis. The diagnostics (\S\ref{sec:diagnostics}) say so in advance --- $1-A$ is near-zero and the neighbourhood aims no better than leaving the input alone, which rules out retrieval here at any depth (\S\ref{app:screenab}, per-layer in \S\ref{app:ab}).


\subsection{External benchmarks: CounterFact}
\label{sec:icl}

The CounterFact benchmark presented here is taken from \citet{meng2022} and run under its own protocol (\S\ref{sec:corpora-external}). Every case asks a different prompt to end in a different object, so no single displacement can serve two cases at once, and the bank is the benchmark's own attribute prompts --- states of other subjects that already carry the requested object. Scored by the benchmark's efficacy measure --- the share of cases where, at the prompt's final position, the first token of the new object outscores that of the true object --- on $60$ held-out cases (Table~\ref{tab:counterfact}, \S\ref{app:counterfact}), \caa\ fails and is indistinguishable from not steering 
($0.233$--$0.250$ against $0.250$). \cccaa\ improves on this, at $0.883$ at layer $20$, and \rgs\ near matches this ($0.767$). This follows from \eqref{eq:endpoint}: every case asking for \textsc{French} wants the same state, so the class holds no variation for a neighbourhood to exploit and the difference between class means is the correct rule. 

The diagnostics predict this. On these test queries retrieval alignment $A$ reads $0.765--0.792$ and the residual edit metric $R$ shows that a single direction leaves $0.991$--$0.992$ of the required displacement, at every layer. \caa\ is predicted to do nothing and does, while the requested class leaves $0.77$--$0.81$, at a $k$ of $16$ against
a class of $18$--$23$: near enough the ceiling that by \eqref{eq:endpoint} the rule being screened is \cccaa\ rather than \rgs\. Coverage barely moves it here ($0.765$--$0.792$ with a state for this case's own destination in
the bank, $0.769$--$0.809$ without): any state carrying the requested object serves, which is
what it means for the class to determine the answer (\S\ref{app:external-screen}).

\subsection{Diagnostics}
\label{sec:screen}

The four quantities of \S\ref{sec:diagnostics} are computed prior to rule fitting. Table~\ref{tab:diagnostics} (\S\ref{app:screens}) scores each against the \caa-minus-\rgs\ margin and against \rgs's own reference error across four task families. Results visualised in Figure\ref{fig:margin_screens} (\S\ref{app:screens}). We evaluate the number of pairs each estimator requires, this is given in detail in \S\ref{app:samples}.

$1-A$, whether the neighbourhood aims where the query has to go, ranks \rgs's error best
($0.65$ mean over families, $0.97$ across the token family) --- it is retrieval's own
field, scored before retrieval runs. It is also the only metric asking for more than the pairs, since
it needs class labels, a bank and hundreds of pairs. Absent a bank, the quantity to read is $R$ (the part of the edit no single vector supplies) at $0.29$ from six pairs, while $D$ (how far the state must move) ranks the same outcome at $0.39$ from the same six. $g$, the measure of whether the required displacements point the same way, \emph{reverses sign} between
families ($+0.27$ on the token corpora, $-0.52$ on the substitution and A/B families) and needs
thousands of pairs where $R$ needs six, so we report it as a component of $R$ and never on its own (\S\ref{sec:failure}). No diagnostic ranks the \emph{margin} reliably, so they are able to say where input-dependence can improve performance, but not by how much.
(\S\ref{app:screendepth}).

\section{Discussion}

A fixed steering rule is insufficient when the required edit depends on the query, and the evaluation on the token corpora quantifies that failure (\S\ref{sec:ladder}). On \textsc{weekday} the required displacements are a permutation of one another and cancel exactly, $g=0.000$ at every captured depth, so \caa\ is not merely imprecise but fits a zero vector; the failure is structural rather than empirical, and the constructions of \S\ref{sec:results-synthetic} reproduce it in a geometry where the correct answer is known in advance. \rgs\ returns a query-specific destination, rectifying the failures of \caa\ whilst differing only in form, not level of supervision. The affine map requires a higher level of supervision in the form of corresponding pairs; where these are not available the method is unusable, and therefore incomparable with other steering methods.

Evaluation on the substitution corpora shows that the advantage is not confined to constructed geometry. \rgs\ lowers reference error in all cross-validated layer-by-corpus experiments across two models. We show the concept-level of the edit decides where in the model to apply it, word-level corpora peaking sharply in the first tenth of the network and clause-level corpora staying flat, a distinction that replicates in a second model (\S\ref{app:models}).

With our proposed diagnostic measures, the data can be used to decide on the more suitable method before either rule is fitted: $R$ orders the families from six unlabelled pairs, and where given a bank, $1-A$ ranks \rgs\'s error at Spearman $0.65$ and reads zero where retrieval has nothing to exploit. Both failures we report were visible to it in advance. On the A/B benchmarks $A$ reads $0.91$--$1.02$ at every layer, so the neighbourhood aims no better than leaving the input alone (\S\ref{sec:failure}). 
Similarly, on CounterFact diagnostics predict \caa\ failure against \rgs\. The screen was built on our own corpora and is read here on a benchmark it was not designed against.

Input-dependent steering may be limited in some ways. \rgs\ costs more computation at steering time by searching a bank that is held in memory, whereas \caa\ simply adds a precomputed vector; we report no timing experiments as the difference in execution time between the methods was negligible in our work. The advantage of \rgs\ rests substantially on coverage (\S\ref{sec:coverage}), which is the intended setting, but is not a universal one. Where coverage is insufficient, however, \rgs\ does not fail arbitrarily: $k$ interpolates between a neighbourhood and the class it is drawn from, and at the class ceiling the two coincide \eqref{eq:endpoint}, so a query whose neighbours carry nothing its class does not already carry is steered by the class-conditional mean.

Taken together, our results suggest the following procedure. Measure $R$ first, at six pairs and no labels; where a bank exists, measure $A$, and treat $1-A\approx0$ as a verdict against retrieval; if given abundant corresponding pairs, fit a map. Where those pairs do not exist the map cannot be fitted, and \rgs\ is the only input-dependent rule available, reading a destination per query from exactly the labels \caa\ already requires. The question a practitioner faces is therefore not which steering rule is better in general, but whether the destination varies inside the class their query requests, and that is measurable before any rule is run.

\section{Conclusion}

A fixed steering rule is insufficient when the embedding geometry and task require edits that depend on the query. We present \rgs\ as an alternative to \caa\ that returns a query-specific destination. \rgs\ rectifies the failures of \caa\ whilst differing only in form, not requiring the additional supervision an affine map demands. In cases where \rgs\ fails, it simply degrades to \cccaa. Not all geometries and tasks benefit from this approach, however. Therefore, we developed  data-derived diagnostic measures that can be used to decide on the more suitable method \textit{a priori}.

\subsubsection*{Reproducibility statement}

All figures and tables regenerate from cached results . Splits, seeds and the full argument list of each run are
recorded in the output JSON. Hyperparameters are selected on validation splits disjoint from
both the bank and the reported score.

\subsubsection*{The use of large language models}

An LLM-based assistant was used to polish prose --- tightening wording and shortening
passages already drafted by the authors. It was not used to generate research ideas, design
experiments, write the analysis code, or interpret results, and all claims and numbers in the
paper were produced and verified by the authors.

\newpage

\bibliographystyle{plainnat}

\appendix
\section*{Appendix}
\addcontentsline{toc}{section}{Appendix}

\section{The affine map}
\label{app:affine}

We evaluate a single global map along side a single global displacement. We define this as, 
\begin{equation}
\delta_{\mathrm{affine}}(h)=\alpha\bigl(Wh+b-h\bigr),
\label{eq:affine}
\end{equation}
with $W\in\mathbb{R}^{d\times d}$ and $b\in\mathbb{R}^{d}$ fitted by ridge regression on the
training pairs $(h_i,r_i)$, minimising $\sum_i\lVert Wh_i+b-r_i\rVert^2+\lambda\lVert
W\rVert_F^2$. The penalty is selected from a logarithmic grid on the validation split, by the same objective is used for the selection of every other rule's
$\alpha$.

It is reported as a ceiling on steering here because it is the most supervised rule
here: it needs each query aligned to the particular state it had to become, which \cccaa\
and \rgs\ does not. Its output is still a function of $h$ alone, which is what places it under the $1/C$ identity ceiling of \S\ref{sec:metrics}
alongside \caa, and what the generated constructions exploit: a task can defeat every global
\emph{displacement} and still be solved by a global linear \emph{map}, so a win over \caa\
that the affine map also achieves is not evidence for an input-conditional destination
(\S\ref{app:synthetic}).




\section{Corpus examples}
\label{app:examples}

How the six corpora of \S\ref{sec:corpora} are built and rendered: the frames, the
construction rules that keep class vocabularies equal, one worked example per corpus, and the
corpus sizes the capture holds.

\begin{figure}[t]
\centering
\includegraphics[width=\linewidth]{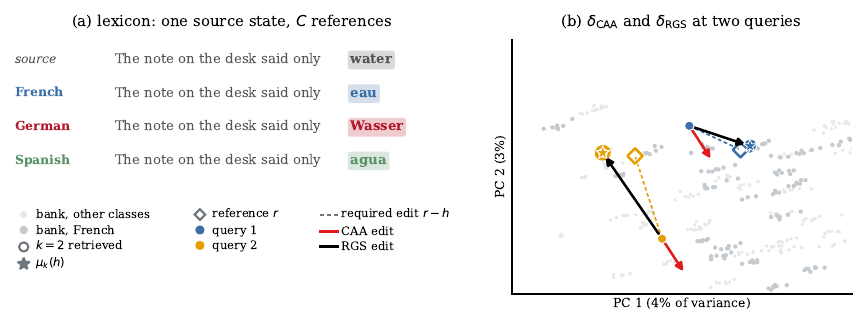}
\caption{Example of a substitution corpus and how the rules act on it. (a) One
concept of \textsc{lexicon} in one frame,
once per class: the prompt is identical across the four renderings and only the highlighted
span differs. The $C$ renderings therefore share a single source state $h$, which is why any rule that
is a function of $h$ alone is capped at $1/C$ on per-item identity (\S\ref{sec:metrics}). (b)
The two rules as operations, on two queries of the same class at layer $0$ of Gemma-2-2B-it.
\caa\ moves both along \emph{the same} vector --- two arrows that are one arrow translated,
and that same vector for every query in the corpus --- while \rgs\ averages the $k$ nearest
bank entries of the requested class and steers to that point, which differs between the two.
Shape says what a mark is and colour which query it belongs to, so a neighbour both queries
retrieved appears as two concentric rings; the dashed line is the step each query required.
The plane carries $7\%$ of the variance and the neighbours were selected in the full space,
so it cannot be read for distance and the retrieved $k$ are ringed rather than left to the
eye. Measured in the full space the queries score $E=0.94$ and $1.01$ under \caa\ against
$0.13$ and $0.96$ under \rgs --- the second is one retrieval does not solve, kept because the
split ordered them that way.}
\label{fig:setting}
\end{figure}

\begin{table}[h]
\centering
\resizebox{\ifdim\width>\linewidth\linewidth\else\width\fi}{!}{%
\begin{tabular}{llclrrr}
\toprule
Corpus & Level & $C$ & Classes & Concepts & Pairs & States \\
\midrule
\textsc{lexicon} & word & $3$ & French, German, Spanish & $25$ & $900$ & $1800$ \\
\textsc{tense} & word & $3$ & past, progressive, third-person & $28$ & $1008$ & $2016$ \\
\textsc{register} & word & $2$ & formal, casual & $28$ & $672$ & $1344$ \\
\textsc{sentiment} & clause & $2$ & positive, negative & $20$ & $480$ & $960$ \\
\textsc{certainty} & clause & $2$ & asserted, hedged & $20$ & $480$ & $960$ \\
\textsc{casing} \emph{(control)} & word & $2$ & upper, title & $28$ & $672$ & $1344$ \\
\bottomrule
\end{tabular}}
\caption{The six substitution corpora. Pairs $=$ concepts $\times$ classes $\times$ $12$ frames, the number of steering problems the corpus poses; states is twice that, a source and a reference for each. Class sizes are equal by construction (\S\ref{app:examples}).}
\label{tab:corpora}
\end{table}

Table~\ref{tab:examples} gives one rendered pair per corpus per class, taken from the first
sentence frame, and Figure~\ref{fig:setting} shows one concept as the model sees it. The source
prompt is identical across the classes of a concept --- only the destination differs --- which
is the property \S\ref{sec:metrics} uses to derive the $1/C$ cap on per-item identity.

\begin{table}[h]
\caption{Rendered source and reference prompts, frame $1$ of $12$. The substituted span is
the last token of each prompt, which is where states are read.}
\label{tab:examples}
\centering\small
\begin{tabular}{llp{0.62\linewidth}}
\toprule
Corpus & Class & Prompt \\
\midrule
\textsc{lexicon} & \emph{source} & The note on the desk said only \textbf{water} \\
 & French & The note on the desk said only \textbf{eau} \\
 & German & The note on the desk said only \textbf{Wasser} \\
 & Spanish & The note on the desk said only \textbf{agua} \\
\addlinespace
\textsc{tense} & \emph{source} & The note on the desk said only \textbf{go} \\
 & past & The note on the desk said only \textbf{went} \\
 & progressive & The note on the desk said only \textbf{going} \\
 & third-person & The note on the desk said only \textbf{goes} \\
\addlinespace
\textsc{register} & \emph{source} & The note on the desk said only \textbf{buy} \\
 & formal & The note on the desk said only \textbf{purchase} \\
 & casual & The note on the desk said only \textbf{grab} \\
\addlinespace
\textsc{casing} & \emph{source} & The note on the desk said only \textbf{water} \\
 & upper & The note on the desk said only \textbf{WATER} \\
 & title & The note on the desk said only \textbf{Water} \\
\addlinespace
\textsc{sentiment} & \emph{source} & She said that \textbf{the meeting ran for an hour} \\
 & positive & She said that \textbf{the meeting was a genuine success} \\
 & negative & She said that \textbf{the meeting was a total failure} \\
\addlinespace
\textsc{certainty} & \emph{source} & She said that \textbf{the report is finished} \\
 & asserted & She said that \textbf{the report is certainly finished} \\
 & hedged & She said that \textbf{the report is possibly finished} \\
\addlinespace
\midrule
weekday & \emph{source} & The budget meeting has been moved to \textbf{Monday} \\
 & next day & The budget meeting has been moved to \textbf{Tuesday} \\
\addlinespace
capitals & \emph{source} & The budget conference will be held in \textbf{France} \\
 & capital & The budget conference will be held in \textbf{Paris} \\
\bottomrule
\end{tabular}
\end{table}

Note the contrast between \textsc{sentiment} and \textsc{certainty}, which is what makes them
a matched pair rather than two clause corpora. \textsc{sentiment} replaces the whole clause,
so its source and reference share only a topic. \textsc{certainty} inserts one hedging adverb
and both renderings end on the same word (\emph{finished}), so the last-token state differs
only in the context that precedes it. The second is the harder test of whether retrieval can
find a propositional attitude without a token-level cue, and \S\ref{sec:level} reports that
retrieval does better on it than on \textsc{sentiment}.

\paragraph{Frames.} Each corpus is rendered in twelve sentence frames chosen to end on the
substituted span and to name nothing from any vocabulary, so the only thing distinguishing a
pair is the span itself. The word-level corpora use frames such as ``The note on the desk
said only \{word\}''. The clause-level corpora need their own, since ``The label on the box
read the meeting was a complete failure'' is not a sentence; they use reporting frames such
as ``She said that \{clause\}'' with deliberately flat verbs, because a frame like
``everyone agreed that'' would put a stance in the frame and the stance is what the class
must carry. Twelve is also what makes a frame-disjoint four-fold split possible.

\paragraph{Construction rules.} Two rules keep the diagnostics honest. Every concept carries
\emph{every} class, so no class has a smaller vocabulary than another and $A$ cannot become
partly a measurement of which concepts happen to be stored. And a concept whose destination
in some class equals the source word is dropped rather than kept with that class removed: a
cognate such as French \emph{table} asks for no displacement at all, and keeping the concept
while dropping one class would leave the classes with different vocabularies.

For the generated constructions, the states are sampled rather than produced by a model:
three isotropic modes with per-mode displacements chosen so that their mean is zero, which
sets $g=0$ exactly and leaves retrieval quality as the only free variable.

\section{Metric definitions in full}
\label{app:metrics}

The definitions behind \S\ref{sec:metrics}, with the estimator choices each one rests on.

\subsection*{State-space metrics}

\paragraph{Reference error.} For query state $h_i$, steered state $h_i'$ and reference $r_i$,
\begin{equation}
E=\frac{1}{n}\sum_i\lVert h_i'-r_i\rVert \;\Big/\; s,
\qquad
s=\frac{1}{n}\sum_i\lVert r_i-h_i\rVert ,
\label{eq:referror}
\end{equation}
so $E=1$ is the score for leaving the input alone, $E=0$ is exact, and $E>1$ means the edit
moved states away from their references. Normalising by the corpus's own mean required step
$s$ makes the scale comparable across corpora and layers whose displacements differ in size.
$E$ is measured against each query's \emph{own} reference rather than against a class
centroid, which matters because distance-to-centroid is minimised exactly by mapping every
query onto that centroid --- the collapse an input-conditional rule exists to avoid.

\paragraph{Per-item identity.} $E$ is an average of norms and so tolerates a rule that lands
every query in the right region without distinguishing between them. Identity asks the
sharper question: for each steered state, is the nearest reference \emph{among all test
references} its own?
\begin{equation}
\mathrm{id}=\frac{1}{n}\sum_i\mathbb{1}\Bigl[\arg\min_j\lVert h_i'-r_j\rVert = i\Bigr].
\label{eq:identity}
\end{equation}
Higher is better. This is the metric the $1/C$ ceiling above binds, for any rule that does
not read the requested class, and Table~\ref{tab:identity} shows the
affine map reaching it exactly on \textsc{casing} ($0.500\pm0.000$) and never exceeding it. Identity is therefore not a
harder version of $E$ but a different question: $E$ asks how close a rule lands, and identity
asks whether it could have distinguished the queries at all.

\paragraph{Destination rate.} The class-level companion to identity, and the readout most
comparable to what the steering literature usually reports. Form one centroid per class
\emph{per side} --- $\mu^{\mathrm{src}}_c$ over that class's sources and
$\mu^{\mathrm{tgt}}_c$ over its references --- and ask whether a steered state's nearest
centroid, among all $2C$, is its own class's target side:
\begin{equation}
\mathrm{dest}=\frac{1}{n}\sum_i\mathbb{1}\Bigl[\operatorname*{arg\,min}_{s\in\{\mathrm{src},
\mathrm{tgt}\},\,c}\bigl\lVert h_i'-\mu^{s}_{c}\bigr\rVert=\mu^{\mathrm{tgt}}_{c_i}\Bigr].
\label{eq:destination}
\end{equation}
Including the source-side centroids is what makes the number start in the right place. A
classifier over destinations alone assigns every state to \emph{some} destination, so an
unsteered query scores well for having moved nowhere; with the source side present it is
nearest its own starting centroid and scores zero.

The rate is a class-level question and is satisfiable by collapse: a rule emitting one point
per class --- which \rgs\ at large $k$ very nearly does --- attains it by construction. It is
therefore reported beside identity \eqref{eq:identity} rather than instead of it, the two
answering "did it reach the right region" and "did it reach the right item". Collapse needs no third metric to
detect it: $E$ is measured against each query's own reference, so a rule mapping every query
to one point scores badly, and identity caps such a rule at $1/C$ by \eqref{eq:identity}.

\subsection*{Output metrics}

State-space results need not survive into text, so we decode continuations under each rule.
The readout that is well defined depends on the level of the concept.

\paragraph{Span match.} For a destination that is a single word, we record whether the
continuation contains it, matched case-sensitively on \textsc{casing} where the case
\emph{is} the task. For a destination that is a clause the question is not well posed: no
rule was going to make the model emit a particular twelve-word clause verbatim, and the
measure reads $0$ for every rule including the ones that changed the output. That is a
property of the metric, not a steering result.

\paragraph{Forced choice.} Defined at both levels. For each query the candidates are that
concept's destinations in \emph{each} class --- available because every concept carries every
class by construction --- scored under the edited model, and we report the share of
probability on the requested one against chance $1/C$. Candidate spans differ in token
length, so scores are mean per-token log-probabilities before normalising. Two properties
make it readable. It needs no external classifier, so no second model's biases enter. And
probe sets are class-balanced, which is load-bearing rather than tidy: a rule with a fixed
preference for one class is right on the queries requesting it and wrong on the rest, so on a
balanced set it scores exactly chance however strong that preference is. On an unbalanced set
the same rule would report the corpus's class prior as though steering had produced it.

\paragraph{Fluency.} Steered text is scored by its mean per-token log-probability under the
\emph{unsteered} model. Read against unsteered text this is misleading, since the unsteered
continuation is that model's own greedy output and sits near the scorer's maximum by
construction, so any intervention starts behind. We therefore report it against a
\textbf{span-swapped} reference: the model's greedy continuation with only the source span
rewritten to the destination. That is the likelihood a \emph{correct} edit attains, and the
number a steered continuation is compared with.

\section{Diagnostics in full}
\label{app:diagnostics}

Before either rule is run we define four quantities that describe a steering task rather than
score a rule on it: how far the state must move (standardised displacement, $D$), whether the
required steps agree (global shared fraction, $g$), how much of the edit is left once the
agreeing part is removed ($R$), and how well a retrieved neighbourhood aims at what each
query requires (retrieval alignment, $1-A$). All are computed from the paired states
$(h_i,r_i)$ and the required edits $d(h_i)=r_i-h_i$ they define; $1-A$ also requires class
labels and a bank to search, which here is the set of references. Every one therefore asks
for more supervision than the rule it screens for, and how many pairs each needs is measured
in Table~\ref{tab:sample-size} (\S\ref{app:samples}), and what each requires in
Table~\ref{tab:diagnostics} (\S\ref{app:screens}).

\paragraph{Retrieval alignment.} Every rule compared here takes the form ``one direction per
query, times one scalar'', and the scalar is selected, so what governs a rule's score is not
the length of its $\alpha=1$ step but how well its directions \emph{align} with $d$. For a
proposed field $u$ the optimal scalar is $\alpha=\langle d,u\rangle/\langle u,u\rangle$ in
closed form, leaving
\begin{equation}
\mathrm{res}(u)=\operatorname{mean}_i\bigl\lVert d_i-\alpha u_i\bigr\rVert \;\big/\;
\operatorname{mean}_i\lVert d_i\rVert ,
\label{eq:align}
\end{equation}
in the units of $E$: $1$ for a field that helps no more than leaving the input alone, $0$ for
one that reproduces $d$ exactly after scaling. Applied to retrieval's own field
$u_i=\mu_k(h_i)-h_i$, minimised over $k$ as the rule's selection would, \eqref{eq:align}
gives the \textbf{retrieval alignment} $A$, reported as $1-A$ to keep $E$'s scale. It asks
whether the bank aims, and aim is all it tests: a neighbourhood pointing correctly but too
far is not penalised, since selecting $\alpha$ would fix that. Unlike the three below it is a
property of the bank as well as of the corpus, and it grants retrieval the best step size and
neighbourhood on the very queries it scores, where the rule must choose both on separate
data, so it bounds what retrieval can achieve rather than predicting what it will.

The \textbf{standardised displacement}
\begin{equation}
D=\sqrt{\mathbb{E}_i\lVert d_i\rVert^2}\;\big/\;\sqrt{\mathbb{E}_j\lVert x_j-\bar{x}\rVert^2}
\label{eq:D}
\end{equation}
is the root-mean-square edit in units of the spread of the pooled states about their centre,
a standardised effect size on the scale Cohen's $d$ is read on. It says nothing about
agreement between edits.

The \textbf{global shared fraction}
\begin{equation}
g=\frac{\lVert\mathbb{E}[d]\rVert}{\mathbb{E}\lVert d\rVert}
\label{eq:g}
\end{equation}
asks whether the required steps point the same way, insensitive to their lengths: $g=1$
describes a task every query asks the same thing of, $g=0$ one whose steps cancel. It is a
directional summary and not a score. It is also expensive to estimate, costing two orders of magnitude more pairs than the other three diagnostics (Table~\ref{tab:sample-size}).

\paragraph{The residual edit.} $R$ is the portion of the edit no single vector supplies, in the
same units as $D$: the root-mean-square displacement about its own mean, over the states'
spread. It is what a per-query rule is left to work with, and the strongest diagnostic
available without labels or a bank.

It is the residual leg of an exact decomposition, $D^{2}=S^{2}+R^{2}$, with $S$ the part a
difference of means does supply, so it can equally be written $R=D\sqrt{1-\lVert\mathbb{E}[d]
\rVert^{2}/\mathbb{E}\lVert d\rVert^{2}}$. It is also written $D\sqrt{1-g^{2}}$, the same
expression with the mean of norms in place of the root-mean-square. The two are not identical
--- $g^{2}$ is the larger by Jensen's inequality, so that form reads the smaller residual ---
and both appear here: the per-task values reported by the library, including every $R$ in the
layerwise tables, are the squared form, while the rank correlations of
Tables~\ref{tab:diagnostics} and~\ref{tab:perfam} are computed from $D\sqrt{1-g^{2}}$, the form
their axis labels can state without a second symbol. The choice does not move them: the two
correlate at $0.999$ or better within every family, and recomputing Table~\ref{tab:perfam} from
the squared form moves no cell by more than $0.01$ ($0.84$ against $0.84$ on the token corpora,
$0.26$ against $0.26$ on the substitution family, $0.40$ against $0.41$ on the A/B benchmarks).

\section{Estimator sample size}
\label{app:samples}

The pair requirement of Table~\ref{tab:diagnostics} is measured rather than assumed, by
subsampling each estimator on a fixed task and comparing against its value on the full
corpus. The targets are the sample sizes at which relative standard error reaches $2\%$.

$R$ reaches it at $6$ pairs and $D$ at $16$, and the elbow does not move across four
generated geometries, which is what makes the sample size one number rather than a per-task
one. $g$ is the contrast: on a task constructed so that the displacements cancel it is
\emph{biased} rather than merely noisy, reading above its true zero until the sample runs
into the thousands ($2048$), because the norm of a mean cannot average its sampling noise
away.

$A$ behaves differently depending on what is being subsampled, and the distinction matters
in practice. Subsampling the \emph{queries} against a fixed bank, it reaches $2\%$ at
$384$--$512$. Subsampling the \emph{pairs}, which shrinks the bank as well, it does not reach
$2\%$ within $512$ on three of four constructions --- the pairs are also the datastore it
searches. This is the sense in which $A$ is cheap to measure and expensive to supply.

Table~\ref{tab:sample-size} collects the requirement for all four. Each is a functional of
the required edits $d_i=r_i-h_i$, so each needs every source linked to the reference it had
to become, and what separates them is how many such links: $R$ at $6$, $D$ at $16$, $1-A$ at
$384$--$512$ queries against a fixed bank, and $g$ not usefully at any sample size we ran.
The correspondence is the same for all four, which is why a diagnostic asks for more
supervision than \rgs\ itself, whatever its sample size.

\begin{table}[h]
\centering
\resizebox{\ifdim\width>\linewidth\linewidth\else\width\fi}{!}{%
\begin{tabular}{llrrrr}
\toprule
Diagnostic & Requires & Pairs to $2\%$ & Fitted slope & Error at $n=32$ & Error at $n=512$ \\
\midrule
$R=D\sqrt{1-g^2}$ & pairs & $6$ & $-0.56$ & $1.0\%$ & $0.2\%$ \\
$D$ & pairs & $16$ & $-0.80$ & $2.0\%$ & $0.2\%$ \\
$g$ & pairs & $2048$ & $-0.52$ & $0.158$ (bias) & $0.039$ (bias) \\
$1-A$, queries subsampled & pairs, labels, bank & $384$--$512$ & --- & --- & --- \\
$1-A$, pairs subsampled & pairs, labels, bank & $>512$ & --- & --- & --- \\
\bottomrule
\end{tabular}}
\caption{The number of corresponding pairs each diagnostic needs. ``Pairs to $2\%$'' is the sample size at which relative standard error reaches $2\%$, read from the recorded power-law fit, and the two right-hand columns give the measured error at two grid points either side of it. A slope near $-0.5$ is the Monte-Carlo rate. $g$ is biased rather than merely noisy on a task whose displacements cancel, which is why its target runs into the thousands. $1-A$ is swept two ways: against a fixed bank it reaches $2\%$ within the grid, while subsampling the pairs shrinks the bank with the sample and 3 of 4 constructions do not reach $2\%$ by $512$.}
\label{tab:sample-size}
\end{table}




\section{Generated constructions}
\label{app:synthetic}

\begin{figure}[p]
\centering
\includegraphics[width=\linewidth]{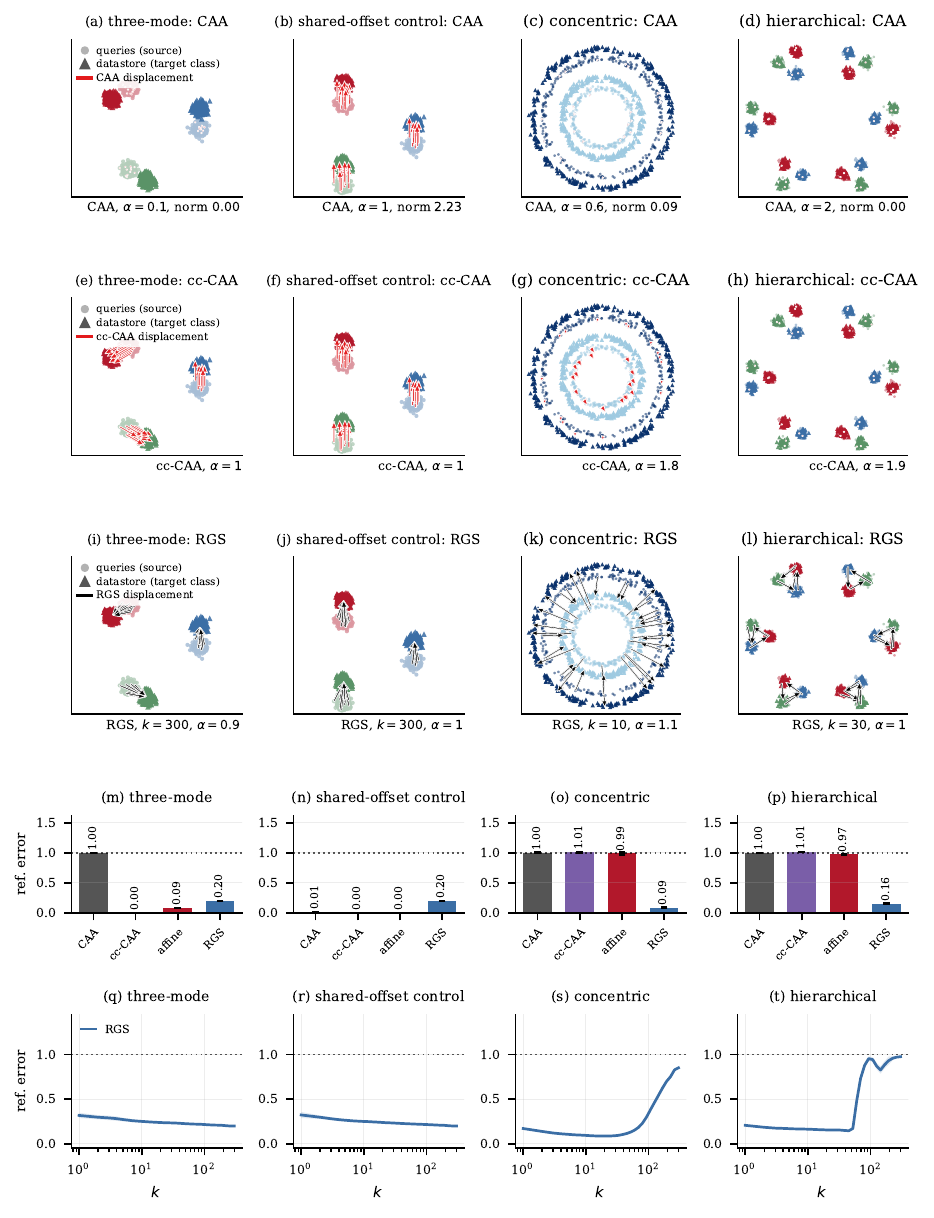}
\caption{Steering rules on constructed geometries. The four
constructions appear in the same order throughout: (a)--(d) draw \caa's edit, (e)--(h)
\cccaa's and (i)--(l) \rgs's, each at the $\alpha$ and $k$ it selected and on one scale, so
only the rule differs between them. \caa is
near zero on all but \textsc{shared offset}. \cccaa\ reproduces the required edit wherever the displacement is constant within a class and collapses to a stub on \textsc{concentric}
and \textsc{hierarchical}, where it cancels within a class as well, while \rgs's arrows turn with the mode, the ring and the cluster. (m)--(p) give Table~\ref{tab:synthetic} as bars and
(q)--(t) sweep $k$, where the rise at large $k$ on the last two constructions is the
neighbourhood mean growing past one ring or cluster.}
\label{fig:synthetic}
\end{figure}

The four constructions of \S\ref{sec:corpora-generated} are positive controls: the geometry
is set rather than measured, so each rule's correct score is known before it is run. All are
two-dimensional with $300$ states per class and five seeds; Figure~\ref{fig:synthetic} draws
all four, including the two of Figure~\ref{fig:method} with their $k$ sweeps, and
Table~\ref{tab:synthetic} reports what each rule scores.

\paragraph{The constructions.} \textsc{three-mode} places three source clusters and rotates
each one's offset, so the required displacements cancel and $g=0$ exactly. \textsc{shared
offset} is the same geometry with one common offset instead, placing $g=1$: it is the control
that isolates $g$ as the variable, since nothing else about the task changes.
\textsc{concentric} puts classes on rings of radius $1$ and $2$ and sends each query radially
outward at its own angle; radial vectors average to zero around a ring, so the mean
displacement vanishes \emph{within} a class as well as across classes, and the two ring
ratios ($2$ and $\tfrac12$) cannot be served by one linear map. \textsc{hierarchical} places
every class inside every cluster with a cluster-specific rotation, so the displacement for a
class depends on which cluster the query is in; the linear part of the required map is shared
across clusters but the intercept is not.

\paragraph{What they establish beyond \S\ref{sec:results-synthetic}.} The selected $k$
separates \rgs\ from the rule it coincides with at $k=|\mathcal{D}_c|$
\eqref{eq:endpoint}: $300$, the class ceiling, where the class mean is the right answer, and
$10$ to $30$ where it is not. And the retrieval alignment $A$ anticipates what \rgs\ then
scores to within $0.020$ on all four constructions, the tightest evidence available that
$1-A$ measures the headroom it is read as measuring.

\begin{table}[h]
\caption{Generated constructions: diagnostics computed before any rule runs, and the
reference error each rule scores, mean $\pm$ s.d.\ over five seeds. $k$ is \rgs's selected
neighbourhood, over the grid $\{1,3,10,30,100,300\}$ and selected per seed; the class holds
$300$ states, so $k=300$ is the class ceiling at which \rgs\ coincides with \cccaa\
\eqref{eq:endpoint}.}
\label{tab:synthetic}
\centering\small
\begin{tabular}{lrrrrrrr}
\toprule
& \multicolumn{2}{c}{Diagnostics} & \multicolumn{4}{c}{Reference error $E$} & \\
\cmidrule(lr){2-3}\cmidrule(lr){4-7}
Construction & $g$ & $A$ & \caa & \cccaa & affine & \rgs & $k$ \\
\midrule
\textsc{three-mode} & $0.000$ & $0.220$ & $1.000$ & $\mathbf{0.000}$ & $0.088$ & $0.200$ & $300$ \\
\textsc{shared offset} & $1.000$ & $0.220$ & $0.012$ & $\mathbf{0.000}$ & $0.000$ & $0.200$ & $300$ \\
\textsc{concentric} & $0.054$ & $0.072$ & $1.001$ & $1.008$ & $0.826$ & $\mathbf{0.081}$ & $10$--$30$ \\
\textsc{hierarchical} & $0.000$ & $0.149$ & $1.000$ & $1.010$ & $0.973$ & $\mathbf{0.155}$ & $30$ \\
\bottomrule
\end{tabular}
\end{table}

\section{The token corpora in full}
\label{app:token}

The supervision ladder of \S\ref{sec:ladder} with its fold spreads: four rules on weekday
advancement and country-to-capital, cross-validated. The two unpaired rules are matched on
class information, so the gap between them is what locality inside the class is worth, and
the affine map is a ceiling rather than an entrant.

\begin{table}[h]
\caption{Reference error on the token corpora, mean $\pm$ s.d.\ over $5$ folds; lower is
better. \caa's weekday vector is zero by identity rather than estimation error, the corpus
permuting one set of states (\S\ref{app:weekday-geometry}).}
\label{tab:token}
\centering\small
\begin{tabular}{lcrr}
\toprule
Rule & Supervision & Weekday & Capitals \\
\midrule
Affine map & paired & $\mathbf{0.195}\pm0.003$ & $\mathbf{0.292}\pm0.008$ \\
\midrule
\cccaa & unpaired & $0.326\pm0.002$ & $0.524\pm0.005$ \\
\rgs\ (target class) & unpaired & $0.307\pm0.004$ & $0.408\pm0.009$ \\
\caa & unpaired & $1.000\pm0.000$ & $0.778\pm0.003$ \\
\bottomrule
\end{tabular}
\end{table}

\section{The weekday corpus's geometry}
\label{app:weekday-geometry}

\begin{figure}[h]
\centering
\includegraphics[width=\linewidth]{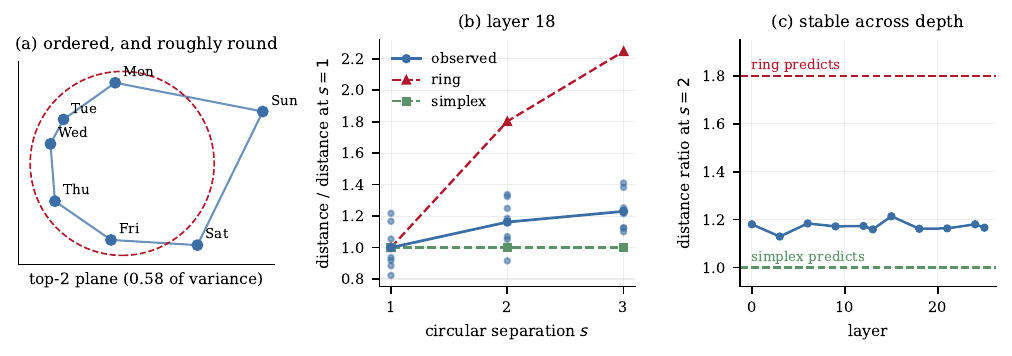}
\caption{The weekday classes at layer $18$ of Gemma-2-2B-it. (a) the seven day centroids in
their top-two plane, in calendar order, with a circle fitted at their mean radius. (b)
distances between centroids by circular separation, normalised by the distance between
neighbours, against what a ring and a simplex of equidistant points predict; each faint point
is one centroid pair. (c) the separation-2 ratio at every captured layer.}
\label{fig:geometry}
\end{figure}

\S\ref{sec:ladder} reports that \caa's weekday vector is zero by identity rather than by
estimation error: the corpus steers each day to the next, so the source and target sets are
the same seven states permuted and their centroids coincide. Figure~\ref{fig:geometry} adds
the shape of the seven classes, because the obvious reading of that shape is wrong in a way
that matters for what a fixed direction could have done.

Projected onto their top two components, which carry $0.58$ of the variance, the day centroids
fall in calendar order around a rough circle (a). A circle would make ``next day'' a rotation,
and a rotation is something a single displacement approximates locally. The measurement
rejects it. On a ring, centroids two and three steps apart sit at $1.80$ and $2.25$ times the
distance between neighbours; here they sit at $1.16$ and $1.23$, far nearer the value $1$ that
equidistant classes would give (b), and the ratio holds within $0.07$ at every layer of the
stack (c).

The seven days are therefore close to mutually equidistant, and ``next day'' is a permutation
of near-equidistant points rather than a step along a curve. That is consistent with the rest
of the token-corpus result: a permutation has no preferred direction, which is why \caa\
scores exactly the no-op value, but it is expressible by a single linear map, which is why the
fitted affine map is the rule that does best here (\S\ref{sec:ladder}).

\section{The substitution corpora at one layer}
\label{app:lexical-direct}

Table~\ref{tab:lexical-direct} is Figure~\ref{fig:lexical-direct-cv} numerically, its panels a bar chart
cannot carry: the score for leaving the input alone, against which every reference error is
read; the affine map, which is the paired ceiling rather than a competitor; and the $\alpha$
and $k$ each rule selected on validation, which are part of the result rather than a footnote
to it. One frame-disjoint split, one layer, no averaging over depth.

One cell is worth locating here: \rgs's selected $k$ is $4$ on every word-level corpus and
$8$ on \textsc{sentiment}, the one clause-level corpus in the table whose identity falls
below the cap --- the same split \S\ref{sec:level} develops across all nine layers.

\begin{table}[h]
\centering
\resizebox{\ifdim\width>\linewidth\linewidth\else\width\fi}{!}{%
\begin{tabular}{llrrrrrrrrrrr}
\toprule
& & & \multicolumn{3}{c}{Reference error $\downarrow$} & \multicolumn{2}{c}{Identity $\uparrow$} & \multicolumn{2}{c}{Destination $\uparrow$} & \multicolumn{3}{c}{Selected}  \\
\cmidrule(lr){4-6}\cmidrule(lr){7-8}\cmidrule(lr){9-10}\cmidrule(lr){11-13}
Corpus & Level & Unsteered & \caa & \rgs & Affine & \caa & \rgs & \caa & \rgs & $\alpha_{\caa}$ & $\alpha_{\rgs}$ & $k$ \\
\midrule
\textsc{lexicon} & word & $0.966$ & $0.826$ & $0.585$ & $0.672$ & $0.32$ & $0.76$ & $0.33$ & $0.67$ & $0.9$ & $0.7$ & $4$ \\
\textsc{tense} & word & $0.990$ & $0.882$ & $0.762$ & $0.826$ & $0.33$ & $0.73$ & $0.33$ & $0.92$ & $0.8$ & $0.5$ & $4$ \\
\textsc{register} & word & $1.004$ & $0.962$ & $0.767$ & $0.779$ & $0.45$ & $0.60$ & $0.50$ & $0.58$ & $0.9$ & $0.6$ & $4$ \\
\textsc{sentiment} & clause & $1.003$ & $0.834$ & $0.573$ & $0.430$ & $0.16$ & $0.13$ & $0.35$ & $0.72$ & $1.0$ & $0.9$ & $8$ \\
\textsc{certainty} & clause & $0.978$ & $0.818$ & $0.565$ & $0.678$ & $0.50$ & $0.82$ & $0.50$ & $0.92$ & $1.0$ & $0.8$ & $4$ \\
\textsc{casing} & word & $0.936$ & $0.768$ & $0.721$ & $0.766$ & $0.50$ & $0.97$ & $0.49$ & $0.78$ & $0.8$ & $0.4$ & $4$ \\
\bottomrule
\end{tabular}}
\caption{\caa\ against \rgs\ on the 6 substitution corpora at layer 13 of Gemma-2-2B-it, on one frame-disjoint test split. $\alpha$ for both rules and $k$ for retrieval are selected on validation and reported unchanged on test. Reference error is in units of each corpus's own required step, so the unsteered column is the score for doing nothing. The affine map is shown as the ceiling rather than as a competitor: it is fitted on matched pairs, which neither unpaired rule receives. Identity is the fraction of steered states whose nearest reference among all test references is its own, capped at $1/C$ for any rule that is a function of the state alone.}
\label{tab:lexical-direct}
\end{table}

\section{The $k$ sweep in full}
\label{app:ksweep}

\begin{figure}[h]
\centering
\includegraphics[width=\linewidth]{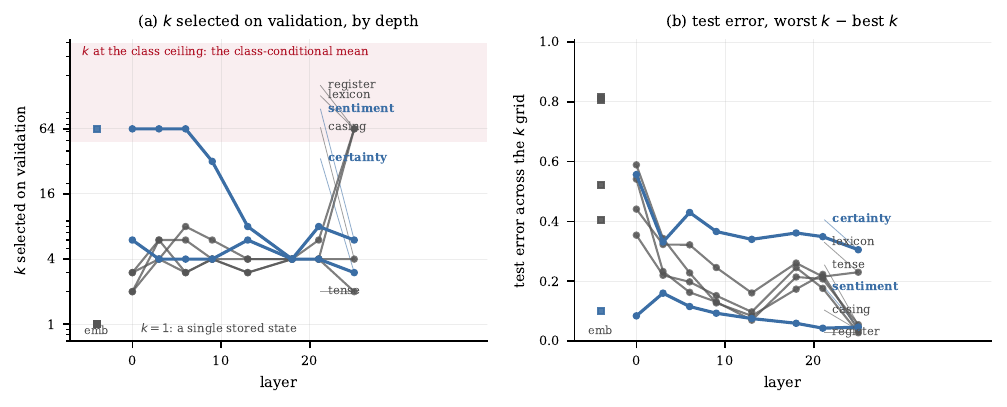}
\caption{Where retrieval is a local rule, and what choosing $k$ is worth. (a) $k$ selected
on validation against depth; the shaded band is the class-ceiling regime, where \rgs\
reduces to \cccaa. (b) the spread of test error across the $k$ grid: how
much the choice was worth.}
\label{fig:kdepth}
\end{figure}

Section~\ref{sec:level} reports the selected $k$ and its fold range. Figure~\ref{fig:kdepth}
gives the underlying curves: where between $k=1$ and the class ceiling each corpus places the
winner at each depth, and how much the choice was worth, measured as the spread of test error
across the whole grid.

The second panel is what licenses reporting $k$ as a regime. On \textsc{sentiment} the curve
is flat everywhere ($0.05$--$0.16$), so no choice of $k$ is worth much and the fold-to-fold
instability of the selection is expected; on the word-level corpora the curve is steepest
exactly where the selection is most local.

Two details of the sweep are worth recording. The sweep grid refines the selection grid and
does not extend past it: the winner is chosen from $\{1,2,4,\dots,64\}$, so a curve drawn to
a larger $k$ would report the class-ceiling regime at a value the selection could never
choose, and a figure and a table describing one run would disagree. And $\alpha$ is
re-selected at every $k$, since a neighbourhood that aims well but reaches short is a
different finding from one that does not aim, and a single step size imposed across the grid
would merge them.

\section{Propagation to text}
\label{app:propagation}

\begin{figure}[t]
\centering
\includegraphics[width=\linewidth]{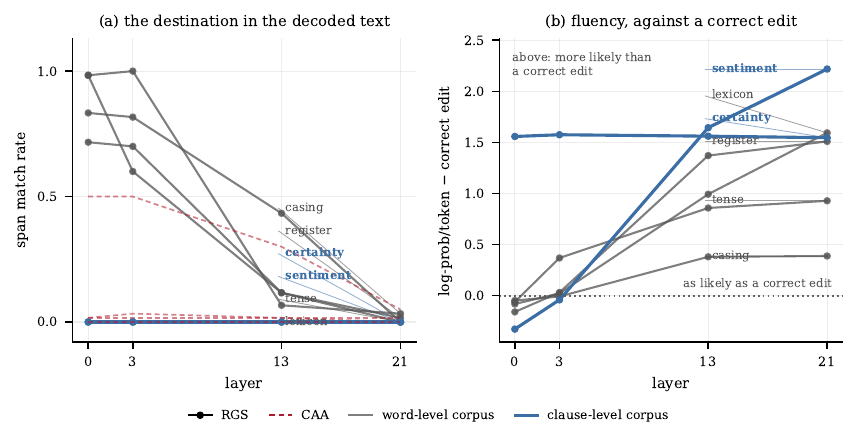}
\caption{The effect of steering on the decoded text ($n=60$). (a) the rate at which the
destination span appears in the decoded continuation; it collapses with depth on every corpus, with \caa\ near
zero throughout except on the \textsc{casing} control. (b) the per-token log-probability of
the same continuations under the \emph{unsteered} model, minus that of a span-swapped
reference --- the model's own greedy continuation with only the source span rewritten --- so
$0$ is the likelihood a correct edit attains. Above the line means the steered text is
\emph{more} likely than a correct edit would be, which past the shallow layers is the same
fact as (a): the edit cost nothing because little was changed. The forced-choice readout of
these runs is quoted in the text.}
\label{fig:propagation}
\end{figure}

Figure~\ref{fig:propagation} gives the two panels behind \S\ref{sec:propagation}, on the
same runs the text quotes.

\textsc{casing} shows why \caa\ sits at chance throughout: it
scores $0.50$ span match there --- which resembles partial success --- and exactly $0.500$
forced choice, since it upper-cases unconditionally, satisfying one class and failing the
other.
Against the span-swapped reference, \rgs's shallow continuations cost what a correct edit
costs, and by layer 21 cost nothing, having changed nothing.

\section{Model depth validation}
\label{app:models}

\begin{figure}[t]
\centering
\includegraphics[width=0.7\linewidth]{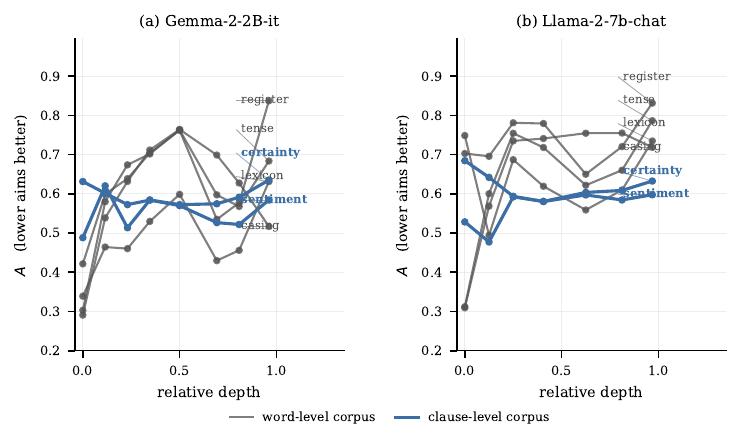}
\caption{Steering in every substitution corpus was evaluated at nine depths of (a) Gemma-2-2B-it and (b) Llama-2-7b-chat. $A$ is plotted against the relative depth in both models.}
\label{fig:models}
\end{figure}

We compare the performance of steering at different model depths across the substitution corpora in both Gemma-2-2B-it and LLama-2-7b-chat, giving the $90$ layer-by-corpus cells \S\ref{sec:level} reports (Figure~\ref{fig:models}). The
comparison is of shape rather than specific levels, since the two models differ in width, depth and tokeniser.

\section{A/B benchmarks, per layer}
\label{app:ab}

Table~\ref{tab:ab} gives the per-layer measurements behind \S\ref{sec:failure}. The margin is in the units of the benchmark's own outcome, the share of probability mass on the
behaviour-exhibiting answer, so positive means retrieval beat the difference of means. $A$ varies by $0.006$ across the six layers of sycophancy and by $0.038$ across corrigibility, while the measured margin varies by $0.098$ and $0.124$ and changes sign on corrigibility.  With this we demonstrate that a diagnostic saturated at the top of its range cannot order outcomes. We are restricted to reading $1-A$ as a verdict on the task and not as a ranking of rule performance over layers.

\begin{table}[h]
\caption{Retrieval alignment $A$ and the measured \caa-minus-\rgs\ margin, by layer, on the
two A/B benchmarks in Llama-2-7b-chat. $A$ is saturated at every layer in both tasks; the margin varies.}
\label{tab:ab}
\centering\small
\begin{tabular}{lrrrr}
\toprule
& \multicolumn{2}{c}{Sycophancy} & \multicolumn{2}{c}{Corrigibility} \\
\cmidrule(lr){2-3}\cmidrule(lr){4-5}
Layer & $A$ & measured margin & $A$ & measured margin \\
\midrule
$5$ & $1.002$ & $+0.022$ & $1.008$ & $+0.006$ \\
$9$ & $1.002$ & $+0.015$ & $1.021$ & $+0.066$ \\
$13$ & $1.008$ & $+0.047$ & $1.026$ & $-0.058$ \\
$17$ & $1.006$ & $+0.111$ & $1.038$ & $-0.040$ \\
$21$ & $1.008$ & $+0.113$ & $1.043$ & $-0.008$ \\
$25$ & $1.008$ & $+0.081$ & $1.046$ & $-0.030$ \\
\bottomrule
\end{tabular}
\end{table}

\begin{table}[h]
\centering
\resizebox{\ifdim\width>\linewidth\linewidth\else\width\fi}{!}{%
\begin{tabular}{llrrrrrr}
\toprule
& & \multicolumn{2}{c}{\caa} & \multicolumn{2}{c}{Task vector} & \multicolumn{2}{c}{\rgs} \\
\cmidrule(lr){3-4}\cmidrule(lr){5-6}\cmidrule(lr){7-8}
Task & Layer & shared & disj. & shared & disj. & shared & disj. \\
\midrule
\textsc{english-french} & 13 & $0.00$ & $0.00$ & $0.00$ & $0.00$ & $0.00$ & $0.00$ \\
\textsc{english-french} & 20 & $1.00$ & $1.00$ & $0.00$ & $0.00$ & $1.00$ & $0.00$ \\
\textsc{country-capital} & 13 & $0.88$ & $0.88$ & $1.00$ & $1.00$ & $0.88$ & $1.00$ \\
\textsc{country-capital} & 20 & $1.00$ & $1.00$ & $0.00$ & $0.00$ & $1.00$ & $0.00$ \\
\textsc{antonym} & 13 & $0.90$ & $0.90$ & $0.90$ & $0.90$ & $0.90$ & $0.90$ \\
\textsc{antonym} & 20 & $0.50$ & $0.50$ & $0.70$ & $0.70$ & $0.80$ & $0.00$ \\
\bottomrule
\end{tabular}}
\caption{Top-1 accuracy on the in-context-learning protocol in Gemma-2-2B-it: the intervention is applied at the last token of a zero-shot query and scored by whether the target word is produced. The unsteered model scores $0.00$ everywhere. All three rules read the same states --- the last token of in-context prompts --- and differ only in using their mean or the $k$ nearest. \emph{shared} lets retrieval's bank hold states for the query's corresponding source--target pair under other demonstration draws; \emph{disjoint} does not. The additive function vector of this literature is \caa\ at these states --- a task-conditioned mean minus the zero-shot mean is a difference of means --- and is reported here under that name; it is the additive part of \citet{todd2023} without the causal head selection, so it is a lower bound on the published method.}
\label{tab:icl}
\end{table}

\section{CounterFact}
\label{app:counterfact}

Cases are drawn balanced across target objects --- $30$ objects, four cases each --- rather
than sampled freely, because an object requested once has only its own attribute prompts in
the bank, and $k$ at the class ceiling is then the whole bank: retrieval would coincide with
\cccaa\ by \eqref{eq:endpoint} before any measurement. With four cases per object the bank
holds $24$ states per class and $k$ has room to be local. Cases are split in half into
fitting and test, $\alpha$ and $k$ are selected on a third of the fitting half, and no rule
sees the query's own prompt.

Two things this run is not. The intervention is a single additive edit at one layer, which is
far weaker than the weight edits CounterFact was built for --- \citet{meng2022} report
efficacy near $1$ --- so the table compares steering rules with each other and not against
model editing. And top-1 accuracy stays low for every rule, so the ordering rests on the
efficacy measure rather than on the argmax.

\begin{table}[h]
\centering
\resizebox{\ifdim\width>\linewidth\linewidth\else\width\fi}{!}{%
\begin{tabular}{lrrrrrr}
\toprule
 & \multicolumn{2}{c}{Layer 6} & \multicolumn{2}{c}{Layer 13} & \multicolumn{2}{c}{Layer 20} \\
\cmidrule(lr){2-3}\cmidrule(lr){4-5}\cmidrule(lr){6-7}
Rule & eff. & top-1 & eff. & top-1 & eff. & top-1 \\
\midrule
Unsteered & $0.250$ & $0.017$ & $0.250$ & $0.017$ & $0.250$ & $0.017$ \\
\caa & $0.250$ & $0.017$ & $0.233$ & $0.017$ & $0.250$ & $0.017$ \\
\cccaa & $0.417$ & $0.000$ & $0.667$ & $0.000$ & $0.883$ & $0.250$ \\
\rgs & $0.350$ & $0.000$ & $0.650$ & $0.017$ & $0.767$ & $0.083$ \\
\bottomrule
\end{tabular}}
\caption{CounterFact \citep{meng2022} in Gemma-2-2B-it, 60 held-out cases. Efficacy is the benchmark's own measure --- the fraction of cases where the edited model prefers the new object to the true one --- and top-1 is the stricter rate at which the new object is the argmax. Every rule reads the same bank, the states of the benchmark's attribute prompts, and $\alpha$ and $k$ are selected on held-out cases. \caa\ is indistinguishable from not steering at every layer; the rule that works is the class-conditional mean, which retrieval matches without beating.}
\label{tab:counterfact}
\end{table}

\section{The screen on the external benchmarks}
\label{app:external-screen}

The diagnostics of \S\ref{sec:diagnostics} read on CounterFact (Table~\ref{tab:external-screen}), on the test queries the accuracies are scored on. The benchmark supplies no paired destination, so we take one from its own material: one of the case's attribute prompts, another subject's prompt that already carries the requested object. It is held out of the bank, and the two conditions ask whether the case's remaining prompts stay in its class (\emph{shared}, the bank the rule is given, \S\ref{app:counterfact}) or go (\emph{disjoint}). All four say the class fixes the destination. $g=0.13$--$0.14$ leaves $R$ within $0.01$ of $D$, so a single vector is left with the whole edit and \caa\ scores $0.233$--$0.250$ against $0.250$ unsteered. $A=0.77$--$0.81$ at a $k$ of $16$ against classes of $18$ to $23$ is the class mean by \eqref{eq:endpoint}, and coverage moves it by under $0.02$ where it moves the in-context tasks by $0.45$: any state carrying the object serves as well as the case's own. The screen orders these rules; it does not size the gap between them.

\begin{table}[h]
\centering
\resizebox{\ifdim\width>\linewidth\linewidth\else\width\fi}{!}{%
\begin{tabular}{lllrrrrr}
\toprule
Layer & Bank & $A$ & $k$ &  $g$ & $D$ & $R$ \\
\midrule
6 & shared & $0.792$ & 16 & $0.132$ & $1.143$ & $1.134$ \\
6 & disjoint & $0.809$ & 16 & $0.132$ & $1.143$ & $1.134$ \\
13 & shared & $0.783$ & 16 & $0.133$ & $1.167$ & $1.157$ \\
13 & disjoint & $0.792$ & 16 & $0.133$ & $1.167$ & $1.157$ \\
20 & shared & $0.765$ & 16 & $0.141$ & $1.171$ & $1.160$ \\
20 & disjoint & $0.769$ & 16 & $0.141$ & $1.171$ & $1.160$ \\
\bottomrule
\end{tabular}}
\caption{Diagnostics computed on the test queries the accuracies in Tables~\ref{tab:icl} and~\ref{tab:counterfact} are scored on. The four quantities of \S\ref{sec:diagnostics}: the retrieval alignment $A$ at its best $k$, the shared fraction $g$, the standardised displacement $D$ and the residual $R$. $g$, $D$ and $R$ read the pairs alone and so do not depend on the bank. The benchmark supplies its own reference state i.e., the case's own attribute prompts. \emph{shared} leaves that case's remaining rows in its class, which is the bank the rule is given; \emph{disjoint} drops them, leaving the same object's entries from other cases.}
\label{tab:external-screen}
\end{table}

\section{The diagnostics across depth}
\label{app:screendepth}
\label{sec:screen-depth}

\begin{figure}[t]
\centering
\includegraphics[width=\linewidth]{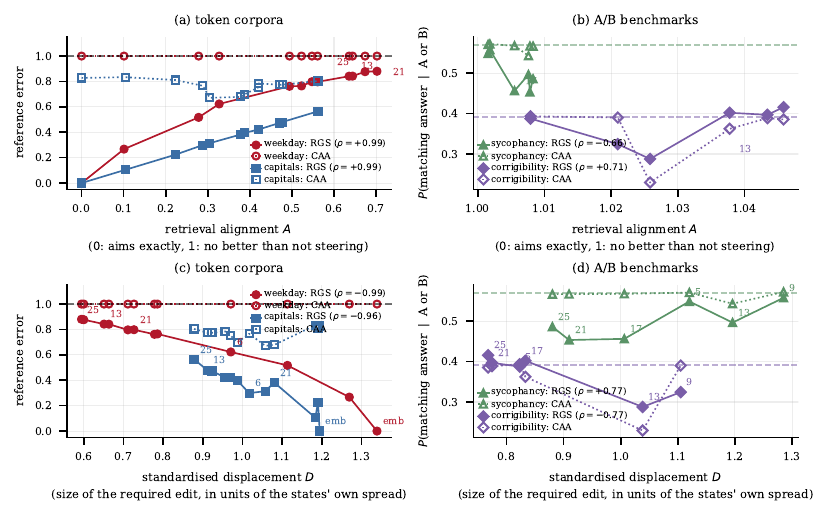}
\caption{Retrieval alignment $A$ (a, b) and standardised displacement $D$ (c, d) against what
the rules score, one point per captured layer of one fixed corpus. Token corpora (a, c) are
read in reference error, the dashed line at $1$ being the no-op score; the A/B benchmarks
(b, d) in the share of probability on the behaviour-exhibiting answer, against each corpus's
own dotted unsteered level. Depth varies the diagnostic without varying the task: weekday
asks the same edit at every layer, but $D$ falls from $1.34$ to $0.59$ as the states' spread
changes.}
\label{fig:outcome}
\end{figure}

\begin{figure}[t]
\centering
\includegraphics[width=\linewidth]{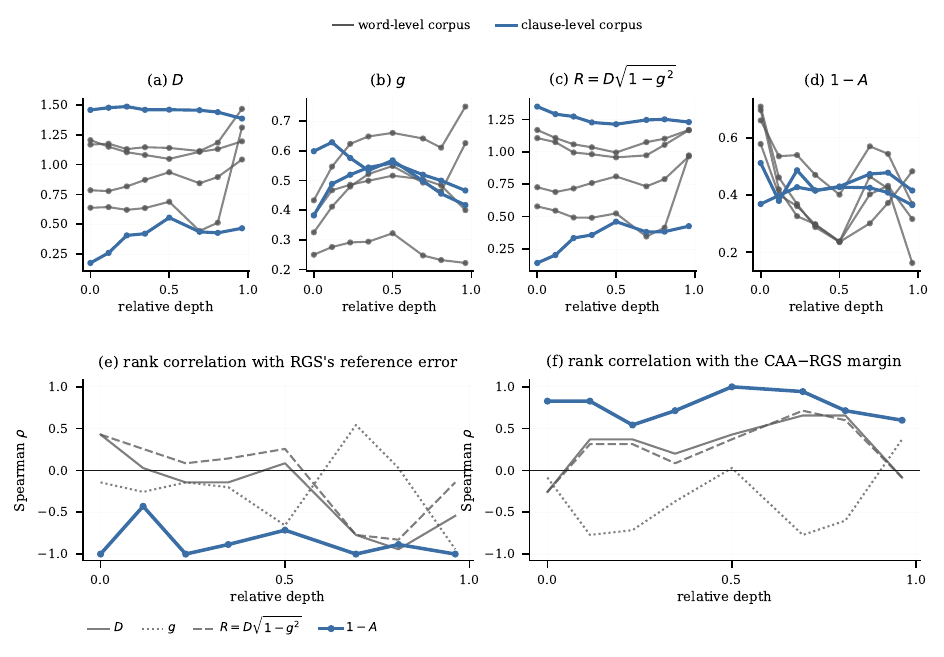}
\caption{The four diagnostics across depth on the six substitution corpora in
Gemma-2-2B-it. (a)--(d) what each one reads, one line per corpus, clause-level corpora in
blue. (e)--(f) whether each still ranks the outcome at that depth, as the rank correlation
over the six corpora within a single layer; $1-A$ is drawn heavier. The embedding is
excluded, as in every lexical depth panel: its states ignore the sentence frame, so the bank
holds each query's own reference and the screen reads a task no later layer poses.}
\label{fig:diagdepth}
\end{figure}

Table~\ref{tab:diagnostics} pools every layer, and a diagnostic can rank corpora against each
other at one depth without tracking one corpus along the stack. Across corpora, at each of
the eight non-embedding layers (Figure~\ref{fig:diagdepth}), only $1-A$ keeps its sign, correlating with \rgs's reference error between $-0.43$ and $-1.00$; $D$ and $R$ rank the outcome as a size diagnostic should
near the input ($\rho=+0.43$ at layer $0$) and reverse to $-0.94$ and $-0.83$ past mid-stack,
and $g$ changes sign twice (\S\ref{app:screens}). Within one corpus, where only the
representation changes (Figure~\ref{fig:outcome}), $A$ tracks the outcome at $\rho=+0.99$ on
both token corpora and is saturated in $[1.00,1.05]$ on the benchmarks, where the
correlations are therefore noise. $D$ reads $-0.99$ and $-0.96$ on the token corpora, the
opposite of its sign across corpora, and takes opposite signs on the two benchmarks, $+0.77$
and $-0.77$: a quantity whose sign depends on which comparison is being made is not one to
screen on alone.

Two cautions belong with the panel. Six corpora give six points per layer, so any single
correlation is noisy; the claim rests on the sign holding across eight layers rather than on
a value at one. And the diagnostics differ in how much they move at all: $1-A$ ranges over
$0.16$--$1.00$ across the corpora and layers drawn, while \textsc{sentiment}'s $D$ varies by
$0.10$ over the whole stack, so a size diagnostic has little with which to separate the
layers of one corpus even where it separates corpora.

\section{What the A/B failure shows about the diagnostics}
\label{app:screenab}
\label{sec:screen-ab}

The benchmark family of \S\ref{sec:failure} is where the diagnostics divide most sharply, and
it is the case that decides which of them to read.

$A$ reads $0.91$--$1.02$ at every layer of both benchmarks, so $1-A$ is zero to within its own
noise and occasionally negative: the class-restricted neighbourhood aims no better than
leaving the input alone, and there is nothing for an input-conditional destination to reach.
That is a family-level verdict rather than a per-layer forecast --- within a range that
narrow, ranking the layers measures noise, and the measured margins do vary in sign across
them (Table~\ref{tab:ab}) --- but it is the verdict that matters, since a practitioner
reading $1-A\approx0$ would not have run retrieval on this corpus at any layer.

$g$ says the opposite, and is wrong. Pooled over items it reads $0.135$ on sycophancy and
$0.458$ on corrigibility, the signature of a task whose required edits disagree. Computed
\emph{within} a fixed answer letter it rises to $0.763$ and $0.848$: one vector describes the
edit almost completely once the letter is held constant. The cancellation described in
\S\ref{sec:failure} suppresses $g$ and makes a format artefact look like input-dependence,
which is the mechanism behind its sign reversal in \S\ref{app:screens}.

The consequence is a rule for which quantity to trust. A low $g$ is necessary for
an input-conditional rule to help and is not sufficient, because a known nuisance variable
can suppress it. $A$, measured on the search the rule actually performs, is not fooled, which
is why $1-A$ and not $g$ is the quantity to read wherever a bank exists. Given a nuisance
factor known in advance, we recommend reporting $g$ conditioned on it beside the pooled
value; the gap between the two, $0.63$ and $0.39$ here, is itself the diagnostic.

\section{Per-family screen correlations}
\label{app:screens}

\begin{figure}[h]
\centering
\includegraphics[width=\linewidth]{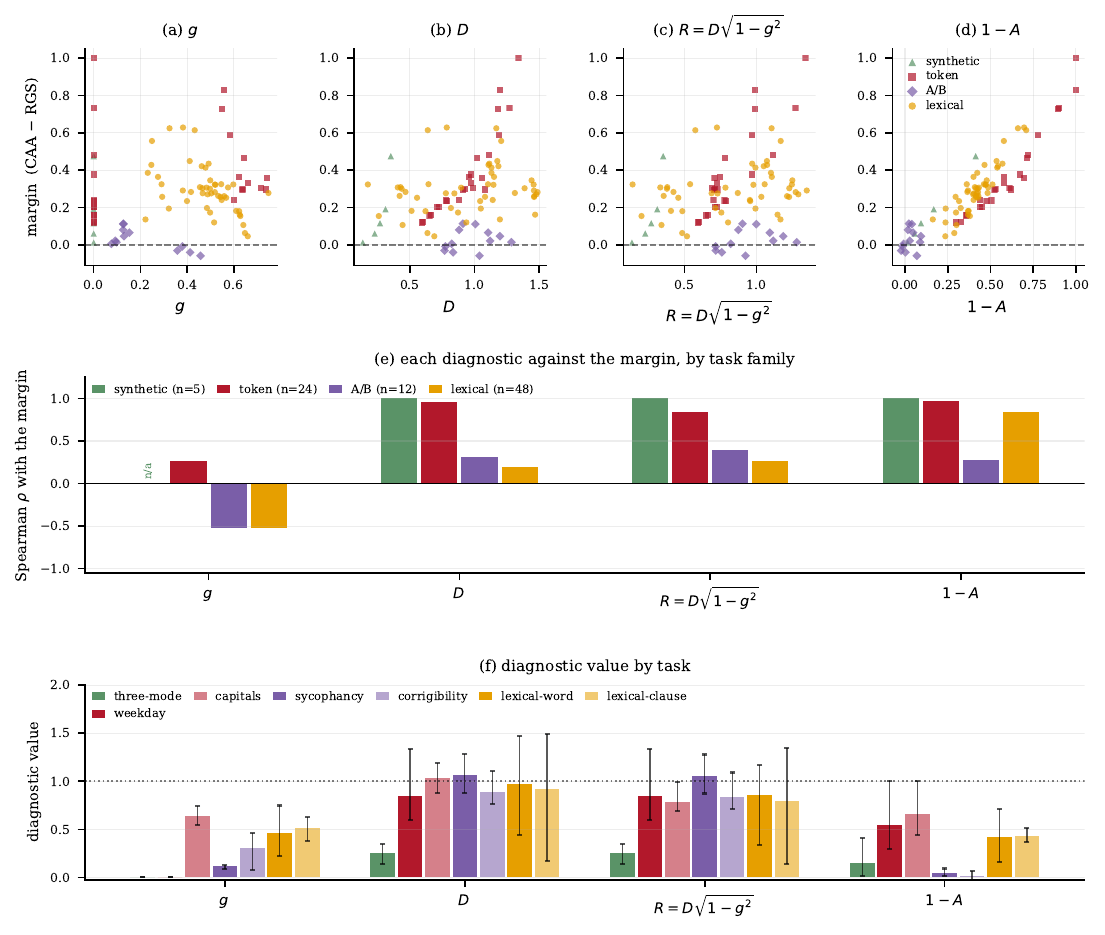}
\caption{The four diagnostics against the measured margin. (a)--(d) scatter by family;
(e)--(h) the correlations of Table~\ref{tab:perfam}; (i)--(l) what each diagnostic
\emph{reads} on each task, which is needed to check that a family's bar is not set by a
quantity that
barely varies there --- the A/B bars for $1-A$ are the case in point, and the reason
\S\ref{sec:failure} reads that family's value rather than its ranking.}
\label{fig:margin_screens}
\end{figure}

Table~\ref{tab:diagnostics} reports the worst family for each diagnostic. Table~\ref{tab:perfam}
gives all four, and Figure~\ref{fig:margin_screens} the scatter each correlation summarises.
Two patterns are only visible here. First, $g$ does not merely rank poorly, it \emph{reverses
sign} between families ($+0.27$ on the token corpora against $-0.52$ on both the A/B and
substitution families), which is why we do not report it alone. Second, the substitution
family is what separates $1-A$ from the size diagnostics: all $48$ of its measurements have a
positive margin, so the margin varies by how much rather than by whether, and a size
diagnostic has nothing to rank it with ($R$ falls to $0.26$) while $1-A$ reaches $0.84$.

The families are never pooled. Their outcomes are on different scales --- a reference error
is divided by its own task's required step, a probability is not divided by anything --- so a
correlation over concatenated points would rank the families against each other rather than
the screens.

\begin{table}[t]
\caption{The four diagnostics: what each asks, what it costs to estimate, and what it ranks.
Rank correlations are Spearman against the measured \caa-minus-\rgs\ margin (worst of four
task families) and against \rgs's own reference error (mean over families). ``Pairs'' is the
sample size at which each estimator reaches $2\%$ relative standard error, measured rather
than assumed (\S\ref{app:samples}). Only $1-A$ needs a datastore, and the pairs used to
estimate it are also the bank it searches. Ordered by what they rank; none ranks the margin
reliably.}
\label{tab:diagnostics}
\centering\small
\resizebox{\ifdim\width>\linewidth\linewidth\else\width\fi}{!}{%
\begin{tabular}{lllrrr}
\toprule
Symbol & Question & Requires & Pairs & $\rightarrow$ margin (worst) &
$\rightarrow$ \rgs\ error (mean) \\
\midrule
$1-A$ & How much can a neighbourhood reach? & pairs, labels, bank & $384$--$512$ &
$\mathbf{0.27}$ & $\mathbf{0.65}$ \\
$R=D\sqrt{1-g^2}$ & How much is left for a per-query rule? & pairs & $\mathbf{6}$ & $0.26$ &
$0.29$ \\
$D$ & How much of the state must move? & pairs & $16$ & $0.20$ & $0.39$ \\
$g$ & Do the required edits point the same way? & pairs & ${\sim}2000$ & $-0.52$ & $0.36$ \\
\bottomrule
\end{tabular}}
\end{table}

\begin{table}[h]
\caption{Spearman correlation with the measured margin, by task family. ``n/a'' marks a
screen that does not vary on that family: the screened synthetic tasks are constructed at
$g=0$ and differ only in floating-point noise.}
\label{tab:perfam}
\centering\small
\begin{tabular}{lrrrr}
\toprule
Screen & Synthetic ($n{=}5$) & Token ($n{=}24$) & A/B ($n{=}12$) & Substitution ($n{=}48$) \\
\midrule
$g$ & n/a & $0.27$ & $-0.52$ & $-0.52$ \\
$D$ & $1.00$ & $0.96$ & $0.31$ & $0.20$ \\
$R=D\sqrt{1-g^2}$ & $1.00$ & $0.84$ & $0.40$ & $0.26$ \\
$1-A$ & $1.00$ & $\mathbf{0.97}$ & $0.27$ & $\mathbf{0.84}$ \\
\bottomrule
\end{tabular}
\end{table}

\section{Cross-validation details}
\label{app:cv}

Folds partition the twelve sentence frames into four disjoint test blocks of three, so every
frame takes exactly one turn as a test frame. Re-seeding a single split instead would give
overlapping test sets and measure resampling rather than the corpus. Validation frames rotate
with the fold, so no frame participates in every selection.

Reported errors are normalised by the whole corpus's required step rather than each fold's,
so that errors from different folds are in one unit. The consequence is that the
unsteered score reads $1.000\pm0.073$ rather than exactly $1$; that spread is a measurement of how much
the frame blocks differ in required step, and is the scale against which every other spread
in the cross-validated tables should be read.

 \end{document}